\PassOptionsToPackage{dvipsnames}{xcolor}
\documentclass[]{fairmeta}

\usepackage{hyperref}
\usepackage{url}
\usepackage{multirow}
\usepackage{booktabs}
\usepackage{graphicx} 
\usepackage{colortbl}
\usepackage{adjustbox}
\usepackage{comment}
\usepackage{makecell}
\usepackage{xspace}
\usepackage{tikz}
\usetikzlibrary{shapes.geometric, arrows.meta, positioning, calc}
\usepackage{amssymb}
\usepackage{placeins}
\usepackage[most]{tcolorbox}

\newcommand{\meta}{\raisebox{-0.2em}{\includegraphics[height=0.9em]{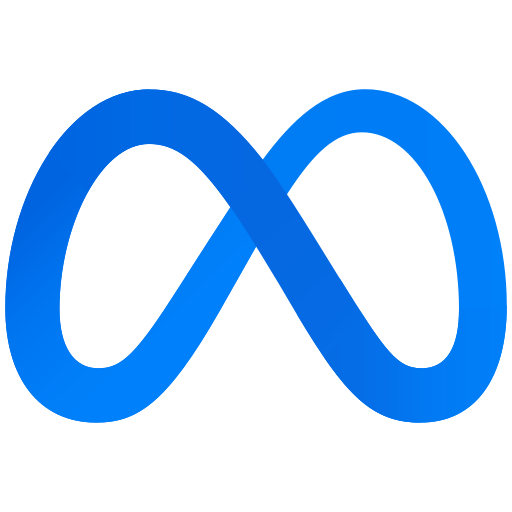}}}
\newcommand{\stanford}{\raisebox{-0.2em}{\includegraphics[height=0.9em]{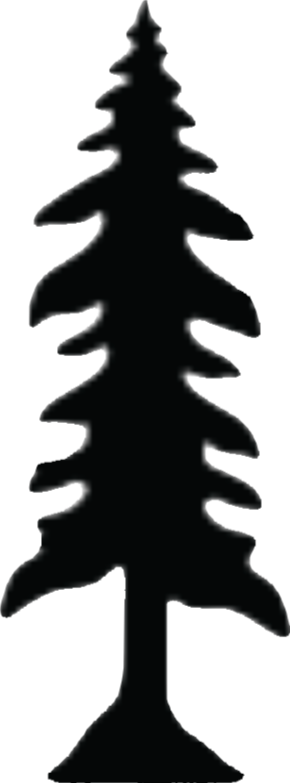}}}
\newcommand{\train}{\includegraphics[height=1.1em]{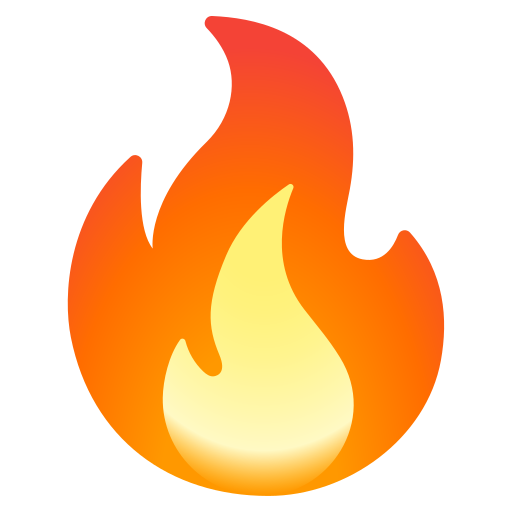}}
\newcommand{\freeze}{\includegraphics[height=1.1em]{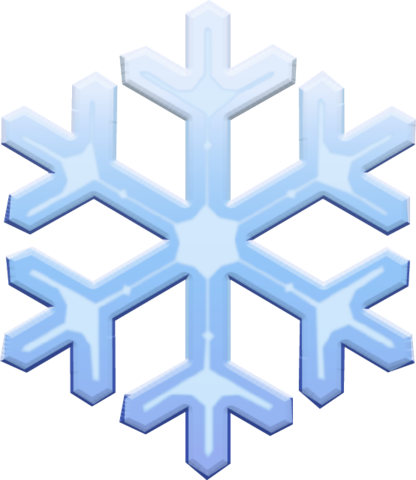}}
\newcommand{\lmmicon}{\includegraphics[height=1.1em]{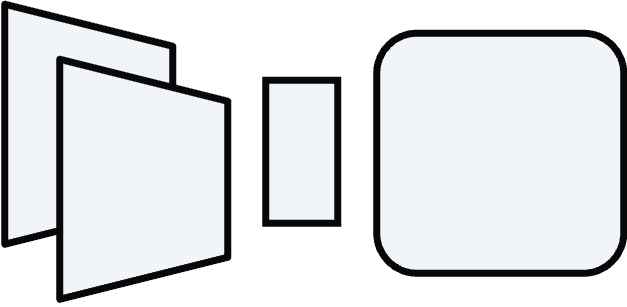}}

\newcommand{\method}{Apollo\xspace}
\newcommand{\benchmark}{ApolloBench\xspace}
\newcommand{\designscale}{Scaling Consistency\xspace}

\newcommand{\finding}[2]{
    \vspace{-0.1cm}
    \begin{tcolorbox}[
        colback=white!90!gray,
        colframe=teal!60!black,
        arc=5pt,
        boxsep=5pt,
        left=10pt,
        right=10pt,
        top=2pt,
        bottom=2pt,
        boxrule=0.8pt,
        drop shadow=gray!0!white,
        enhanced jigsaw
    ]
    \vspace{-0.1cm}
        \paragraph{\textbf{\textit{Finding #1:}}} #2
    \vspace{-0.1cm}
    \end{tcolorbox}
    \vspace{-0.1cm}
}

\definecolor{metaBlue}{RGB}{24,119,242}
\definecolor{metaBlueLight}{RGB}{220,230,242}

\tikzstyle{database} = [
    cylinder,
    shape border rotate=90,
    aspect=0.25,
    draw,
    fill=green!30!yellow!10,
    minimum height=3em,
    minimum width=3em,
    text centered,
    text width=10em
]
\tikzstyle{process} = [
    rectangle,
    draw,
    fill=cyan!30!green!20!yellow!20,
    text centered,
    rounded corners,
    minimum height=3em,
    text width=11em
]
\tikzstyle{decision} = [
    diamond,
    draw,
    fill=cyan!30!green!20!yellow!20,
    aspect=2,
    text centered,
    minimum width=6em,
    minimum height=3em,
    inner sep=0pt,
    text width=8em
]
\tikzstyle{arrow} = [thick,->,>=stealth]

\usepackage[toc,page,header]{appendix}
\usepackage{minitoc}

\title{\method: 
An Exploration of Video Understanding in Large Multimodal Models}

\newcommand{\om}{\ensuremath{\omega}}

\author[\meta, \stanford, \om]{Orr Zohar}
\author[\stanford]{Xiaohan Wang}
\author[\stanford]{Yann Dubois}
\author[ \meta ]{Nikhil Mehta}
\author[\meta]{Tong Xiao}
\author[\meta]{Philippe Hansen-Estruch}
\author[\meta]{Licheng Yu}
\author[\meta]{Xiaofang Wang}
\author[\meta]{Felix Juefei-Xu}
\author[\meta]{Ning Zhang}
\author[\stanford]{Serena Yeung-Levy}
\author[\meta]{Xide Xia}

\affiliation[\meta]{Meta GenAI}
\affiliation[\stanford]{Stanford University}

\contribution[\om]{Work done at Meta}

\abstract{
Despite the rapid integration of video perception capabilities into Large Multimodal Models (LMMs), the underlying mechanisms driving their video understanding remain poorly understood.
Consequently, many design decisions in this domain are made without proper justification or analysis. 
The high computational cost of training and evaluating such models, coupled with limited open research, hinders the development of video-LMMs.
To address this, we present a comprehensive study that helps uncover what effectively drives video understanding in LMMs.

We begin by critically examining the primary contributors to the high computational requirements associated with video-LMM research and discover \textit{\designscale}, wherein design and training decisions made on smaller models and datasets (up to a critical size) effectively transfer to larger models. 
Leveraging these insights, we explored many video-specific aspects of video-LMMs, including video sampling, architectures, data composition, training schedules, and more. For example, we demonstrated that fps sampling during training is vastly preferable to uniform frame sampling and which vision encoders are the best for video representation.

Guided by these findings, we introduce \textbf{\method}, a state-of-the-art family of LMMs that achieve superior performance across different model sizes. Our models can perceive hour-long videos efficiently, with \method-$3$B outperforming most existing $7$B models with an impressive 55.1 on LongVideoBench. \method-$7$B is state-of-the-art compared to 7B LMMs with a $70.9$ on MLVU, and $63.3$ on Video-MME. 
}

\date{\today}
\correspondence{Orr Zohar at \email{orrzohar@stanford.edu}, Xiaohan Wang at \email{xhanwang@stanford.edu}}

\metadata[Project Page]{\url{https://apollo-lmms.github.io}}

\begin{document}
\maketitle
\doparttoc 
\faketableofcontents 
\section{Introduction}
\label{section:intro}

Despite the rapid advancements in language and image-language modeling~\citep{chinchilla, brown2020language, yang2024qwen2, liu2024visual,alayrac2022flamingo,idefics3,openai2024gpt4o}, the development of video Large Multimodal Models (video-LMMs) has not kept pace. 
Videos provide a rich, dynamic information source, capturing nuanced temporal and spatial features beyond the reach of static images.
However, video-LMMs remain under-explored, hampered by unique challenges: notably higher computational demands and a broader, more complex design space compared to their image-based counterparts~\citep{video_chat,llama-vid,oryx,aria,xu2024pllava}.

Many fundamental questions about video-LMM design remain unanswered:
How should videos be sampled? 
Which vision encoders yield optimal representations? 
What are the best practices for resampling video tokens? 
Early approaches primarily extended image-LMMs directly~\citep{xu2024slowfast,image_grid,freeva,longva} or with
video-specific fine-tuning~\citep{video_chat, video_llama,maaz2023video}. 
Recent methods introduced diverse design choices, such as longer context windows~\citep{longva}, multi-modality mixing~\citep{llava,li2024llava}, agent workflows~\citep{videoagent}, self-training~\citep{videostar}, and more. Despite these efforts, the impact of these design decisions on video-LMM performance is poorly understood. This lack of systematic investigation motivates our study.

To overcome the computational challenges of training video-LMMs, we explore whether design decisions from smaller models correlate effectively with larger ones. Traditional scaling laws~\citep{chinchilla} predict model performance based on size, but apply to models trained from scratch and require training multiple models to predict performance. Scaling laws have also been observed in LMM pretraining~\citep{mm_scaling_laws,mm_scaling}. Since LMMs integrate multiple pre-trained components, it's uncertain if these laws hold. 
By relaxing scaling laws, our experiments reveal that design choices made with smaller LMMs transfer to larger ones, a phenomenon we term \textbf{\designscale} (Sec.~\ref{section:scaling-laws}).

Utilizing these insights, we conduct an extensive study across the video-LMM design space, addressing essential aspects of video-language modeling, such as video sampling and encoding methods, token resampling and integration strategies, and data compositions (Sec.~\ref{sec:arch} \&~\ref{sec:training}).
For instance, we discover that frames-per-second video sampling significantly outperforms standard uniform sampling used in previous works~\citep{oryx, kangaroo}. We also find which vision encoder combinations are the most robust and that the Perceiver Resampler~\citep{perceiver} outperforms average pooling. 
When studying the numerous benchmarks available, we discovered a large portion of the performance improvements are driven primarily via language modeling and, therefore, curate \benchmark, which significantly reduces evaluation time while improving assessment quality (Sec.~\ref{sec:bench}).

\begin{figure*}[t]
     \centering
    \includegraphics[width=1\textwidth]{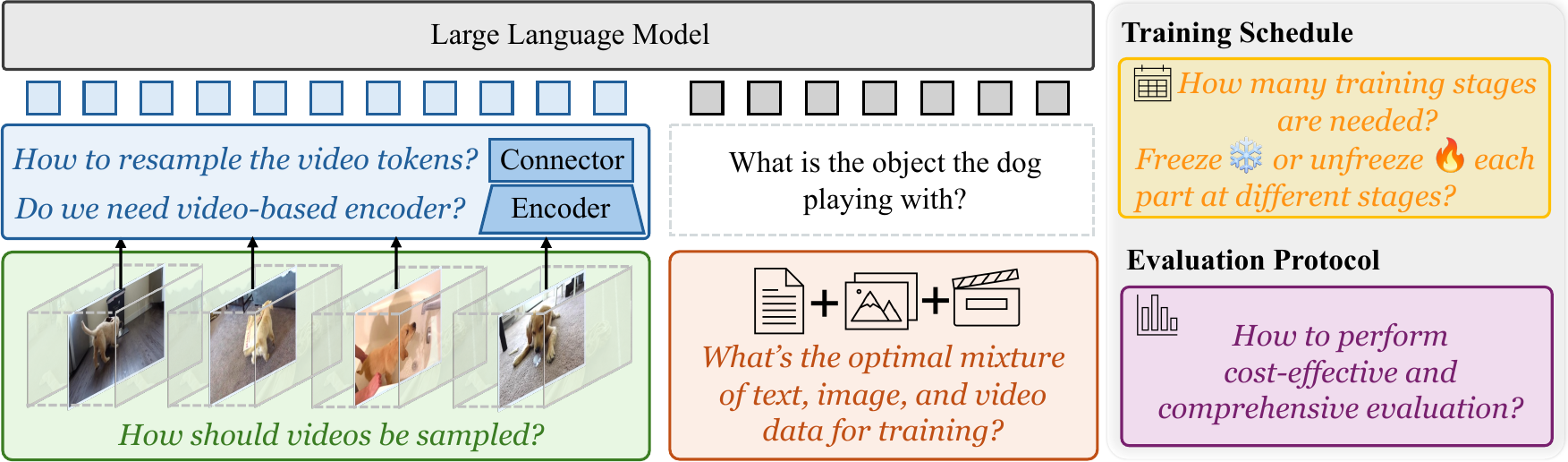}\vspace{-0.1in} 
    \caption{\textbf{\method exploration.} Schematic illustrating our comprehensive exploration of video-specific design choices; critically evaluating the existing conceptions in the field, from video sampling and model architecture to training schedules and data compositions. 
    For example, we found that the SigLIP encoder is the best single encoder for video-LMMs but can be combined with additional encoders to improve temporal perception, and that keeping a $\sim10\%$ text data during fine-tuning is critical for video understanding performance. More insights can be found in Sec.~\ref{sec:arch} \&~\ref{sec:training}.\vspace{-0.15in} 
    }
\label{fig:method}
\end{figure*}

Building upon our findings, we introduce \method, a family of state-of-the-art LMMs capable of comprehending hour-long videos. \method models exhibit strong performance across various scales. Notably, \method-$3$B surpasses most existing $7$B models, achieving scores of $58.4~(+12.8)$ on Video-MME (w/o sub.), $68.7~(+6.9)$ on MLVU, and $62.7~(+14.1)$ on \benchmark. \method-$7$B attains impressive scores of $61.2~(+0.6)$ on Video-MME (w/o sub.), $70.9~(+5.4)$ on MLVU, and $66.3~(+2.4)$ on \benchmark, making it competitive with $30$B models. 
\textbf{Our contributions are as follows:}

\begin{enumerate}
    \item We conduct a systematic exploration of the video modeling design space for Large Multimodal Models, uncovering critical factors that drive performance and providing actionable insights for future research.
    \item We identify \designscale, where design decisions effective for smaller LMMs and datasets are transferred effectively to larger ones, reducing computational costs and enabling efficient experimentation.
    \item We address evaluation inefficiencies by curating \benchmark, a subset of existing benchmarks that cuts evaluation time by $41\times$ while offering detailed insights into temporal reasoning and perception tasks.
    \item We introduce \method, a family of LMMs that achieves state-of-the-art results across video understanding multiple benchmarks. Notably, \method-$3$B surpasses nearly all $7$B models, while \method-$7$B variant is state-of-the-art among models with less than $30$B parameters. 
\end{enumerate}

In Sec~\ref{sec:bench}, we analyze the state of video benchmarks and introduce \benchmark. In Sec.~\ref{section:scaling-laws}, we show how one can relax traditional scaling laws for computational savings. In Sec.~\ref{sec:arch}, we explore the architecture design space. In Sec.~\ref{sec:training}, we investigate different training protocols and data mixtures. Finally, in Sec.~\ref{sec:apollo}, we present \method, a state-of-the-art family of video-LMMs.

\begin{figure*}[t] 
\centering 
\includegraphics[width=1\linewidth]{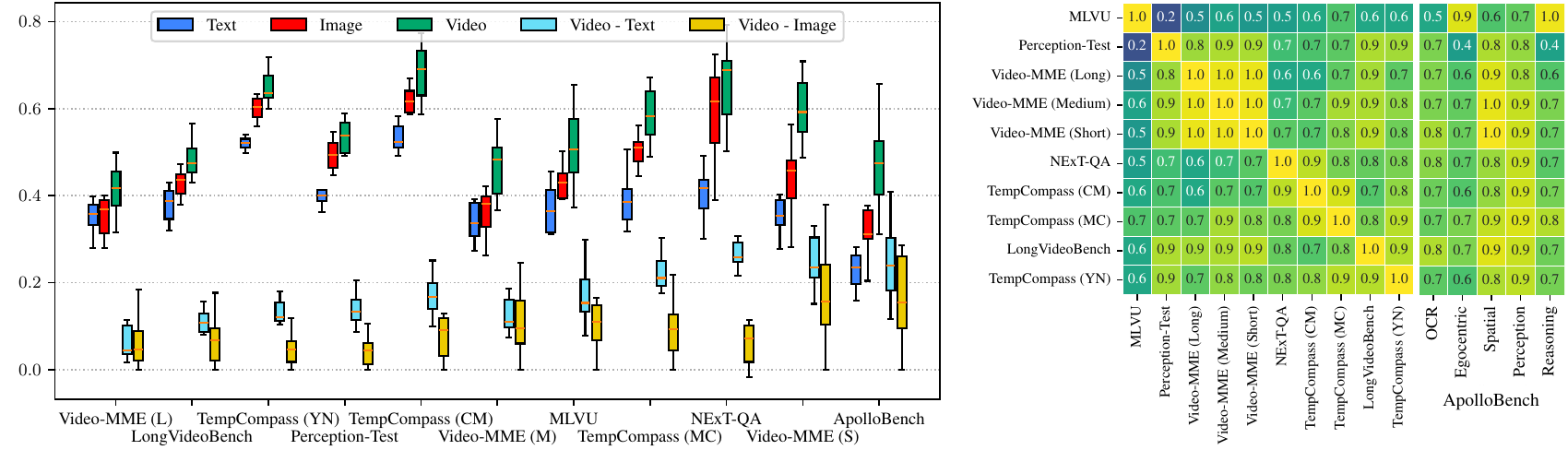}\vspace{0.07in}
\caption{\label{fig:benchmark_analysis} \textbf{Benchmark Analysis.} \textbf{(Left)} Accuracy of the open-source LMMs on various video question-answering benchmarks when provided with different input modalities: full video (green bars), a single frame from the video (red bars), and text-only input without any visual content (blue bars). The light blue shaded areas represent the difference in accuracy between video and text inputs, highlighting the extent to which video perception enhances performance over text comprehension alone. The yellow shaded areas indicate the difference between video and image inputs, quantifying the additional benefit of temporal information from videos compared to static images. 
\textbf{(Right)} The correlation matrix shows the redundancy among benchmarks by illustrating the correlation coefficients between model performances on different benchmarks. Each cell in the matrix represents how closely the two benchmarks are related in terms of model performance. Our proposed benchmark, \benchmark, is highly correlated with all tested benchmarks, suggesting that it offers an equally effective evaluation while being more computationally efficient.}
\vspace{0.07in}
\end{figure*}

\section{How effective are existing video question-answering benchmarks?}
\label{sec:bench}

The rapid advancement of video Large Multimodal Models (video-LMMs) has spurred the creation of numerous video question-answering benchmarks, including Video-MME, MLVU, LongVideoBench, and others~\citep{videomme,longvideobench,mlvu,perceptiontest, li2024videovista, lvbench, temporalbench}. 
While this proliferation enables comprehensive evaluation, it also introduces significant resource intensiveness and redundancy. For example, evaluating a 3B-parameter model on these benchmarks requires 184 A100 GPU hours.
In this section, we first analyze the quality of existing benchmarks (Sec.~\ref{sec:bench:quality}), their redundancy (Sec.~\ref{sec:bench:redundancy}), 
and introduce \benchmark (Sec.~\ref{sec:bench:apollo}) by building on these insights.

\subsection{Evaluating benchmark quality}
\label{sec:bench:quality}

What drives video benchmark performance is not known. As shown by~\citet{goyal2017making}, some image question-answering benchmarks are largely driven by text comprehension rather than image perception.~\citet{mmstar} further showed that data leakage in either the LLM or LMM training stage may be further contaminating evaluation in image question-answering benchmarks. 
To evaluate the state of video question answering benchmarks, we evaluated ten open-source LMMs on several benchmarks: Video-MME~\citep{videomme}, TempCompass~\citep{tempcompass}, LongVideoBench~\citep{longvideobench}, MLVU~\citep{mlvu}, NExTQA~\citep{nextqa}, and PerceptionTest~\citep{perceptiontest}—under three different settings:

\begin{itemize} 
\item \textbf{Video}: Models prompted with video input using their standard video sampling. Green in Fig.~\ref{fig:benchmark_analysis}, left.
\item \textbf{Image}: Models are provided only the center frame of each video. Red in Fig.~\ref{fig:benchmark_analysis}, left.
\item \textbf{Text}: Models are prompted with only the original question, without any visual input. Blue in Fig.~\ref{fig:benchmark_analysis}, left.
\end{itemize}

As illustrated in Fig.~\ref{fig:benchmark_analysis}, left, a significant portion of existing benchmarks are answered solely through text comprehension alone (blue boxplots) or only using the center frame (red boxplots), indicating that LMMs do not rely on video perception in a large portion of existing benchmarks.
We sorted the benchmarks by the difference between the Video and Text performance (light blue). A benchmark relies more and more on its video perception capabilities when this bar is high. 
When examining Fig.~\ref{fig:benchmark_analysis}, left, it is apparent that as videos get longer, the reliance on video perception decreases (compare Video-MME S/M/L). 
To evaluate how much of the benchmarks require video input to answer the question, we also plot the difference between the Video and Image performance (yellow). Some benchmarks can almost be entirely solved using a single frame. 
For example, in line with~\citet{buch2022revisiting}, we find that NExTQA is solved using a single frame. Perception-test also behaves similarly. Finally, when studying Fig.~\ref{fig:benchmark_analysis}, left, a high variance in the box plot is desired as this indicates more highly discriminative benchmarks. Among all the existing benchmarks, Video-MME (Short), MLVU, and TempCompass emerge as the top performers.

\subsection{Redundancy in existing benchmarks}
\label{sec:bench:redundancy}

To evaluate the redundancy in video question answering benchmarks, we evaluated ten open-source LMMs on several benchmarks: Video-MME~\citep{videomme}, TempCompass\citep{tempcompass}, LongVideoBench~\citep{longvideobench}, MLVU~\citep{mlvu}, NExTQA~\citep{nextqa}, and PerceptionTest~\citep{perceptiontest}. We then calculated the correlation of each of the benchmarks to each other, the result of which can be seen in Fig.~\ref{fig:benchmark_analysis}, right. 
Our analysis revealed significant redundancy among benchmarks, as evidenced by the block-diagonal correlation matrix, where we can identify groups of benchmarks that are highly correlated.

To evaluate the effect of different question types and video durations, we also evaluated the correlations between video duration groups. 
We find that the performance of models on short and long videos within Video-MME~\citep{videomme} exhibits an $R^2=0.94$, see App. Fig.~\ref{sup:fig:eval:videomme}, while in LongVideoBench,  $R^2>0.92$ between all duration groups. 
To assess the effect of question format, we studied the TempCompass~\citep{tempcompass} dataset, which has different question formats (multiple-choice, yes/no, caption matching, and caption generation), and found that they are also highly correlated ($R^2>0.8$), indicating that varying question types do not significantly diversify the evaluation (see App. Fig.~\ref{sup:fig:eval:tempcompass}).

\subsection{Introducing \benchmark}
\label{sec:bench:apollo}
Motivated by these insights, we set out to curate a more effective and efficient benchmark suite called \benchmark. We focused on multiple-choice questions to eliminate the need for external tools like ChatGPT, ensuring a consistent and cost-effective evaluation process~\citep{freeva}. 

We filtered out questions that could be correctly answered by more than 50\% of the models with either text or image inputs, removing questions that do not require video perception (see Fig.~\ref{fig:benchmark_analysis}, left, \benchmark). Subsequently, we identified five broad temporal perception categories: Temporal OCR, Egocentric, Spatial, Perception, and Reasoning. Questions were then manually categorized into each one of these categories.  
We selected the top $400$ questions from these categories that exhibited the most discrimination between models via entropy and manually verified each one to validate the correctness of the selected questions. 
Evaluating on \benchmark is $41\times$ faster while being highly correlated with existing benchmarks (see Fig.~\ref{fig:benchmark_analysis}, right) and more influenced by video perception (Fig.~\ref{fig:benchmark_analysis}, left). For more details, see App. Sec.~\ref{app:sec:benchmark_analysis} and App. Fig.~\ref{fig:benchmark_creation_flowchart}.

\section{\designscale: How small can you go during model design?}
\label{section:scaling-laws}

Developing Large Multimodal Models (LMMs) poses significant computational challenges, especially when training on extensive datasets with billion-parameter models. To make the research process more efficient, it is essential to determine whether smaller LMMs and datasets can reliably inform design decisions for larger ones.
Traditional scaling laws require training multiple models of varying sizes for each design decision to derive how performance scales with model size. However, in the context of LMMs, which typically utilize multiple pre-trained components (e.g., vision encoders, language models), scaling each component individually is impractical due to the lack of availability of such components and the immense computational resources required. As such, we set to relax these scaling laws and instead reason about correlation or transfer of design decisions between models of different sizes. 

\begin{figure*}[t]
     \centering
         \includegraphics[width=1\textwidth]{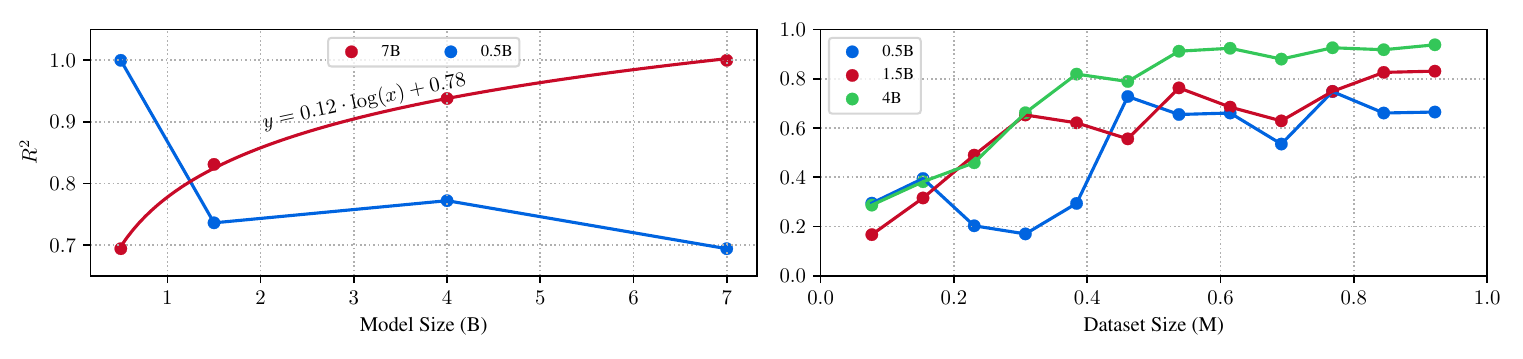}
     \caption{
     \textbf{\designscale.} We discover \designscale, where design decisions made with smaller models on smaller datasets carry over to larger models on larger datasets. \textbf{(Left)} $R^2$ values of
     {\color{BrickRed}\textbf{7B}} and {\color{NavyBlue}\textbf{0.5B}} versus other LLM sizes show an increasing correlation with larger LLM sizes for the 7B model. The same trend is not seen in the $0.5$B model. 
     Interestingly, while the Qwen$1.5$-$4$B model variants have lower/similar performance to their smaller Qwen$2-1.5$B counterparts, the correlation to larger models is still higher (See App. Fig.~\ref{sup:fig:scaling_consistency}).
     \textbf{(Right)} $R^2$ of $0.5/1.5/4$B models to $7$B vs dataset size. $R^2$ to larger datasets starts to plateau at around $500$K samples.
     \label{fig:scaling}
     }\vspace{0.07in}
\end{figure*}

This section investigates the correlation between design decisions made on LMMs of different sizes. Specifically, we selected $21$ model variations encompassing various design aspects such as architecture, video sampling methods, training strategies, and data mixtures. Each variation was trained using four different Large Language Models (LLMs): Qwen$2$-$0.5$B, Qwen$2$-$1.5$B, Qwen$1.5$-$4$B, and Qwen$2$-$7$B~\citep{bai2023qwen,yang2024qwen2}, resulting in a total of $84$ models. We then analyzed the correlation ($R^2$) between the performance of these models (see App. Fig.~\ref{sup:fig:scaling_consistency}).
Our findings reveal that design decisions on models of a critical size ($\sim 2-4$B) correlate highly ($R^2 > 0.9$) with those on larger models, a phenomenon we term \designscale (see Fig.~\ref{fig:scaling}). For instance, the $R^2$ between the $4$B and $7$B models is $0.938$, indicating a strong predictive relationship. Please refer to the App. Sec.~\ref{app:sec:scaling} for a detailed analysis.

Scaling laws typically require training models of various sizes to study performance trends. However, due to limited availability and high computational cost, scaling laws are rarely applied to LMMs.
In contrast, \designscale demonstrates that design decisions made on moderately sized models ($\sim 2-4$B) and datasets transfer reliably to larger models, even across different model families. This allows researchers to make informed design choices without extensive scaling studies. Our primary goal is to show that design decisions transfer reliably, reducing computational burden and accelerating research.

\paragraph{Large Language Model size.}
In Fig.~\ref{fig:scaling}, left, we plot the $R^2$ values between models of various sizes and the 7B LLM model variant. The correlation with the $7$B LLM increases approximately log-linearly with the size of the smaller LLMs and generalizes between model families. This behavior is not observed with smaller models, e.g., $0.5$B, where $R^2$ immediately drops below $0.8$, and no log-linear behavior can be observed. This reinforces the existence of a critical model size ($\sim 2-4$B) where design decisions transfer reliably—a phenomenon we term \designscale. 
\designscale seems to generalize between model families, as a mix of Qwen1.5 and Qwen2 models were utilized in this study. For example, while the Qwen2-1.5B and Qwen1.5-4B model variants had similar performance, the 4B Qwen1.5-4B was still more correlated than the 1.5B model. 
Please refer to the App. Sec.~\ref{app:sec:scaling} for a comprehensive analysis.

\paragraph{Impact of dataset size.}
We examined the impact of dataset size on model performance by training models using the same data mixture but varying the dataset size from $75$K to $1$M samples. The results are shown in Fig.~\ref{fig:scaling}, right, where the correlation of the $0.5/1.5/4$B models trained on varying datasets sized to $7$B trained on the full dataset can be seen as a function of dataset size. 
Focusing on the 4B LLM variant, we observed that the correlation ($R^2$) with larger models plateaus around $\sim500K$ samples, indicating that increasing the dataset size beyond this point yields diminishing returns in terms of informing design decisions. In contrast, smaller models (e.g., 0.5B and 1.5B) exhibited less consistent behavior, with their $R^2$ values fluctuating more across different dataset sizes. This suggests that a dataset size of approximately 500K samples is sufficient for moderately sized models (2–4 billion parameters) to reliably transfer design insights to larger models.
\vspace{0.05in}

\finding{1}{We discover \designscale, where design decisions can be made on smaller models and datasets and transfer reliably to larger ones.} 
\section{Exploring the video-LMM design space: what influences effective model design?}
\label{sec:arch}
In this section, we analyze key architectural design choices shaping the performance of Large Multimodal Models (LMMs) in video-language tasks.
We focus on four critical aspects:
\textbf{(I) Video sampling} (Sec.~\ref{sec:arch:sampling}) where we compare uniform and fps video sampling and evaluate the effect tokens and frames per second have on downstream performance. 
\textbf{(II) Video representation} (Sec.~\ref{sec:arch:encoders}) where we explore how image and video encoders impact video representation and show which encoder and encoder pairs lead to the best performance.
\textbf{(III) Video token resampling} (Sec.~\ref{sec:arch:resamplers}) where we test different visual token resamplers. 
\textbf{(IV) Video token integration} (Sec.~\ref{sec:arch:integration}) where we examine various strategies to integrate the visual token into the text tokens.

Using \designscale, we opted to perform the following exploration using Qwen$2.5$ $3$B~\citep{yang2024qwen2} and trained on a dataset of $750$K samples. 
As demonstrated in Sec.~\ref{section:scaling-laws}, these findings exhibit a strong correlation ($R^2>0.9$) with results on larger models and across different model families. Unless stated otherwise, a Perceiver Resampler~\citep{perceiver} was employed, with $16$ tokens per frame at a frame rate of $2$ fps. The dual encoders used were InternVideo$2$~\citep{internvideo2} and SigLIP-SO$400$M~\citep{siglip}. When training on images, images were duplicated before being encoded by the video encoders for fully integrated encoding, as we found it to be slightly more performant with fewer parameters and complexity (see App. Sec.~\ref{sec:training:unified_split}). This is in line with~\cite{video_llava}.

\begin{figure*}[t]
\centering
\includegraphics[width=\linewidth]{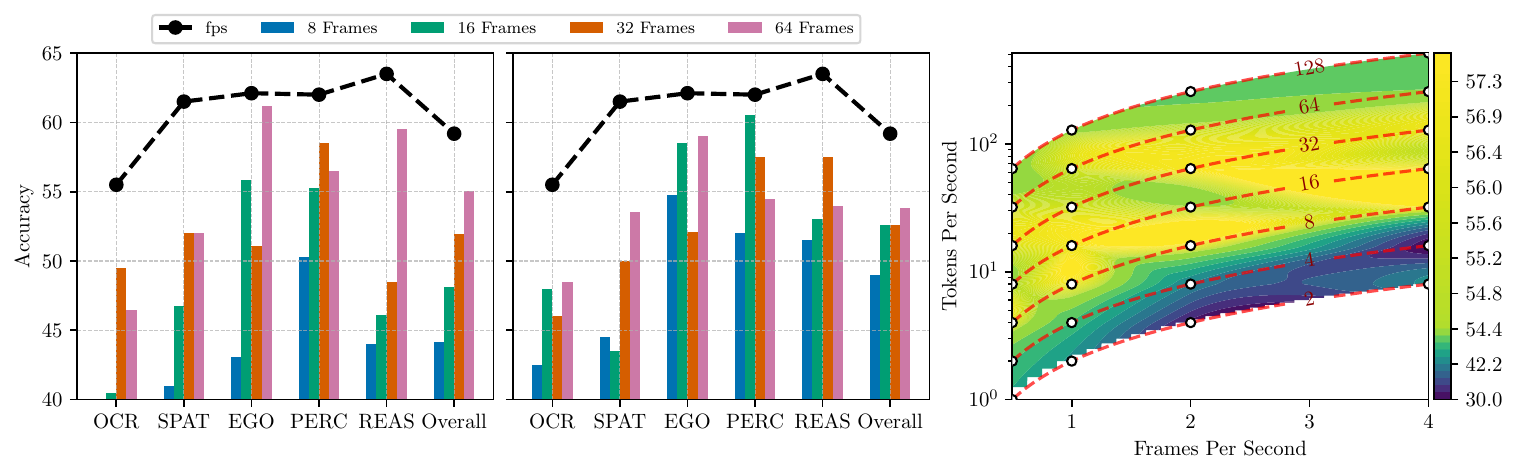}\vspace{-0.1in}
\caption{\textbf{Video sampling.} We compare different sampling strategies and their effect on performance.
\textbf{(Left)} Models were trained and tested using uniform sampling. Increasing the number of frames improves overall performance but does not reach fps sampling performance.
\textbf{(Middle)} Models trained with uniform sampling but tested with fps sampling. Differences in performance are not explained by the number of frames sampled at test time.
\textbf{(Right)} Analysis of the effect of frames per second (fps) and tokens per second (tps) on overall performance. The dotted red lines ({\color{BrickRed}\textbf{-~-}}) indicate the tokens per frame. For a per-metric breakdown, please see App. Fig.~\ref{sup:fig:full_sampling}.\label{fig:sampling} 
}
\end{figure*}

\begin{figure*}[t]
\centering
\includegraphics[width=1\linewidth]{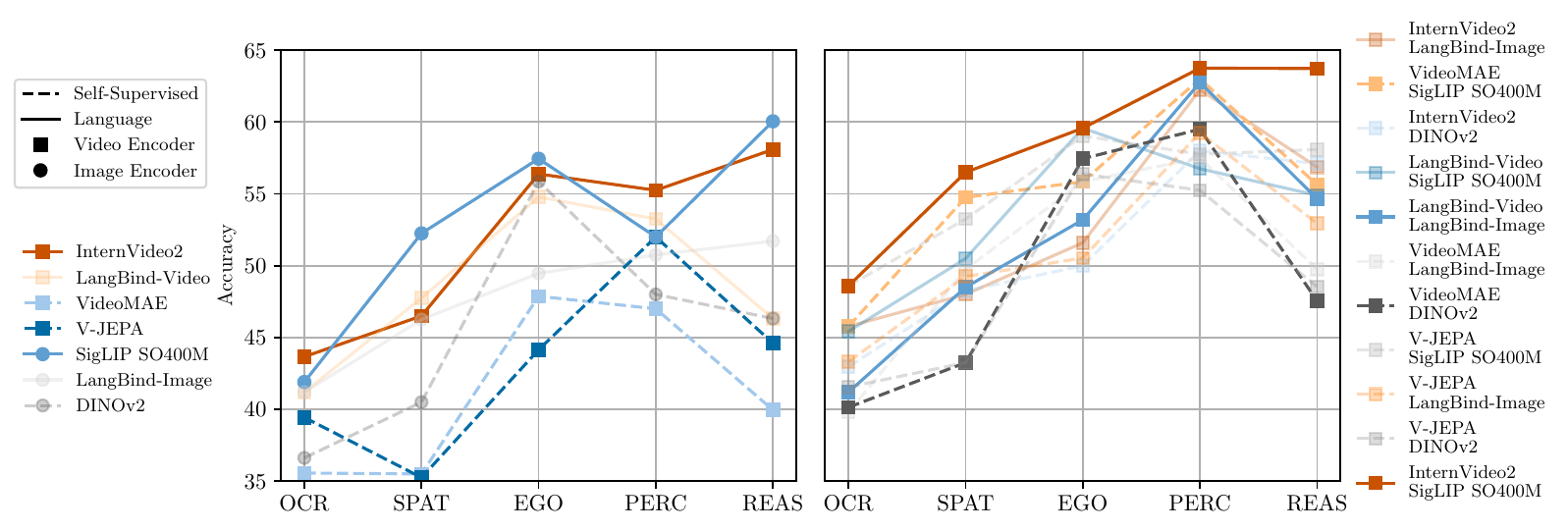}
\caption{\textbf{Vision encoders.} In our study, we tested InternVideo2~\citep{internvideo2}, LanguageBind-Image/Video~\citep{languagebind},V-JEPA~\citep{vjepa}, Video-MAE~\citep{videomae}, SigLIP-SO$400$M~\citep{siglip}, and DINOv2~\citep{dinov2}, and their combinations. \textbf{(Left)} SigLIP-SO-$400$M emerges as the best overall among single encoders. We also find that image encoders underperform in temporal perception compared to video encoders. \textbf{(Right)} Performance of dual-encoder configurations. Language-supervised encoders outperformed their self-supervised counterparts. Combining InternVideo2 and SigLIP-SO-$400$M leads to the best overall performance. \label{fig:vision_encoders}
}
\end{figure*}

\subsection{Video sampling}
\label{sec:arch:sampling}
Videos can be sampled in many ways, from uniform sampling - uniformly sampling $N$ frames from the video~\citep{video_chat, video_llava,chat_univi,zhang2024llavanextvideo}, to fps sampling - sampling at a set number of frames per second. While many recent methods have preferred fps sampling~\citep{oryx,llava}, they default to uniform sampling when video durations exceed their frame sampling capacity (usually $\sim64$). 
The maximum frame capacity is typically constrained due to the memory requirement at the vision encoder and or the LLMs context window.

Uniform frame sampling enables simplified training because the effective `vision batch size' (i.e., the number of frames that need to be encoded) remains constant. However, training video-LMMs with uniform frame sampling means that the time difference between concurrent frames changes with each video, effectively setting a different `video speed' in every iteration. 
As a result, when uniformly sampling $N$ frames from videos of varying lengths, the effective playback speed represented in the sampled frames changes. For a shorter video, $N$ uniformly sampled frames represent a slower playback (more frames per second of actual content), while for a longer video, those same $N$ frames represent a faster playback. This will likely hamper the LMMs' capability to learn about the speed of objects in videos. Meanwhile, methods employing fps sampling must either limit the maximum video duration or the maximum number of frames (above which, they default to uniform frame sampling) in training or suffer from similar issues as in uniform sampling. 
An alternate approach is to sample `video clips' of $N$ frames at a set fps (or duration) and, when reaching the maximum token count, space these out instead. Here, rather than uniformly spacing out the sampled video frames, the $N$ frames encoded by the video encoder maintain the same effective fps, and only frames of concurrent `clips' are spaced out.
Methods that utilize video encoders~\citep{video_chat,video_llava}, where multiple frames are encoded together, should use such frame sampling as video encoders are typically trained at a constant fps~\citep{vjepa,internvideo2,languagebind,videomae}. 

To evaluate the effect of fps vs. uniform sampling, we trained four models that, while training, we uniformly sampled $8, 16, 32$, or $64$ frames. To test whether performance differences are due to the different frame sampling at test or train time, we evaluated these models with uniform and fps sampling. The results of this experiment can be seen in Fig.~\ref{fig:sampling}, left and middle. We found that uniform frame sampling consistently underperformed compared to fps sampling, Fig.~\ref{fig:sampling}, left. As can be seen, this performance gap is not due to the different number of frames sampled at test time, Fig.~\ref{fig:sampling}, middle.
Therefore, we conclude that the uniform frame sampling of videos causes this performance gap during training.

\vspace{0.05in}
\finding{2}{fps sampling is preferable over uniform sampling during model training and inference.}
\vspace{0.05in}

\begin{figure}[t]
\centering
\includegraphics[width=\linewidth]{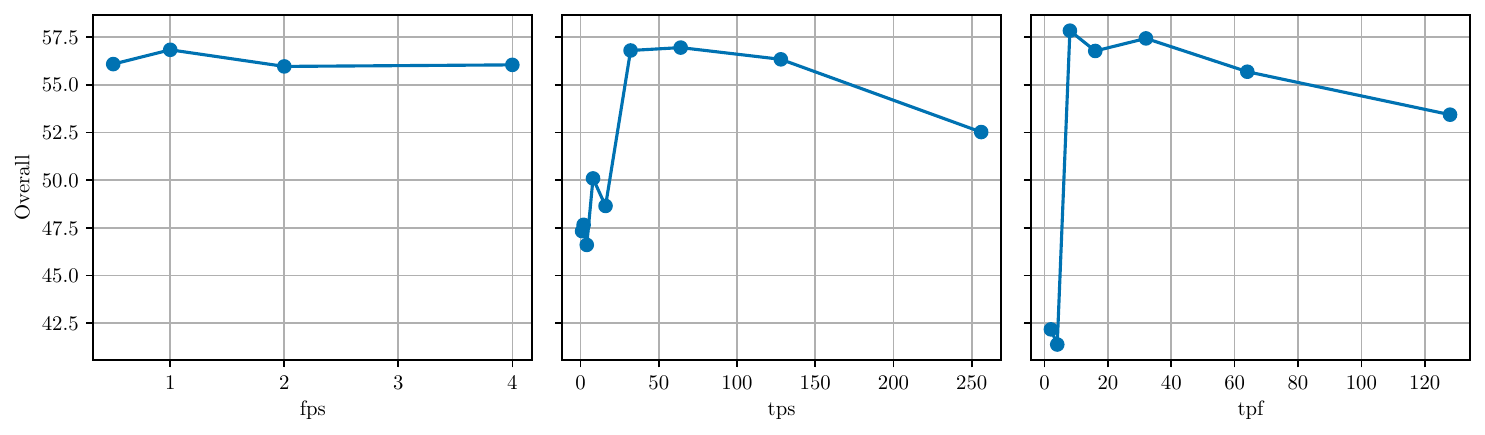}\vspace{-0.1in}
\caption{\label{sup:fig:fps_sampling} \textbf{Video fps sampling analysis.}
Comparison of overall performance across different parameters. The first plot illustrates the impact of frames per second (fps) on performance, while the second and third plots show performance trends with varying tokens per second (tps) and tokens per frame (tpf), respectively.}
\end{figure}

When training at a constant fps, the tokens per second (tps) can also be varied using the token resampler. We investigate how varying fps and tps affect the LMM's ability to comprehend videos.
As can be seen in Fig.~\ref{fig:sampling}, right, 
there appears to be a tradeoff between tps and fps, balancing short and long video performance, with $8$–$32$ tokens per frame achieving strong performance at different fps.
Surprisingly, as can be seen in the App. Fig.~\ref{sup:fig:fps_sampling}, we found little dependence on fps, with both tokens per frame (tpf) and tps being more determinate. 
In concurrent work,~\citet{context_video_llm} reached similar conclusions but required more tokens per frame ($\sim49$) to achieve performance saturation, likely as they utilized only image encoder and average pooling, which is less compressible. They also utilized uniform frame sampling, which may also affect this comparison.

\vspace{0.05in}
\finding{3}{There is a trade-off between tps and fps, with $8$-$32$ tokens per frame being optimal.}
\vspace{0.05in}

Some methods employ active frame selection strategies, using the initial query to guide frame sampling~\citep{videoagent,goldfish,videotree}, and were not included in this study. Note that these would require frame resampling at every conversational turn.

\subsection{Video representation}
\label{sec:arch:encoders}

Training effective video encoders is challenging due to the high memory requirements for processing large video datasets and the comparatively low quality of available supervision. 
While early approaches predominantly used dedicated video encoders~\citep{video_chat,mvbench,video_llava}, recent developments favor image encoders instead~\citep{kangaroo,llava,oryx}. 
This shift arises because image encoders, although lacking temporal integration, still produce higher-quality representations that the LLM can readily leverage.
Another possibility is that in this approach, image and video datasets can be fully integrated, possibly benefiting from image-video transfer and allowing the utilization of the much larger, more diverse, and more efficient image instruction tuning datasets~\citep{llava,longva}. 

Multiple studies have conducted extensive investigations into visual representation within image-LMMs~\citep{nveagle,cambrian1}.~\citet{idefics2} found that SigLIP outperformed even much larger encoders, such as EVA-CLIP-5B.~\citet{griffon} showed that input image resolution influences performance more than token count, which may have influenced~\citet{idefics2}'s ablation.~\citet{visualtokenizer} compared encoders trained with supervision and found where each is preferable. However, whether image or video encoders are preferable for video-LMMs and what influences their performance is unclear. As such, we set out to find what drives good video representation in LMMs. We trained LMMs with several image and video encoders and their combinations and evaluated how this design decision impacted the final model performance. 
Our study includes diverse language- and self-supervised video/image encoders:
\begin{itemize}
    \item \textbf{InternVideo2}~\citep{internvideo2}: trained in two stages: (1) unmasked video token reconstruction, (2) crossmodal contrastive learning aligning video with audio, speech, and text. Encodes four frames.
    \item \textbf{LanguageBind-Video v1.5}~\citep{languagebind}: initialized with an OpenCLIP model, trained contrastively with a frozen text encoder. Encodes eight frames.
    \item \textbf{VideoMAE}~\citep{videomae}: trained through self-supervised learning by masking video patches with a reconstruction loss. Encodes sixteen frames.
    \item \textbf{V-JEPA}~\citep{vjepa}: trained through self-supervised learning by predicting masked spatio-temporal regions in a learned latent representation space. Encodes sixteen frames.
    \item \textbf{SigLIP-SO400M}~\citep{siglip}: a shape-optimized model trained using a sigmoid loss function for language-image pre-training.
    \item \textbf{LanguageBind-Image}~\citep{languagebind}: one of the OpenCLIP image encoders and is not further tuned. 
    \item \textbf{DINOv2}~\citep{dinov2}: trained using a self-supervised teacher-student framework where the teacher guides the student to produce consistent representations across different image views.
\end{itemize}

As seen in Fig.~\ref{fig:vision_encoders}, left, language-supervised encoders consistently outperform self-supervised encoders, in line with observations in prior work~\citep{nveagle}. In the single-encoder setups, SigLIP-SO$400$M had the best performance compared to all image/video encoders, demonstrating that video encoders must be improved to replace image encoders. Video encoders outperform image encoders only on Temporal Perception, indicating that LLMs struggle with fine-grained temporal integration (e.g., estimating speed and direction of movement).

\vspace{0.05in}
\finding{4}{SigLIP-SO$400$M is the best single encoder for video-LMMs.}
\vspace{0.05in}

We hypothesize that combining video and image encoders could offset their limitations, where image encoders do not encode temporal information, while video encoders have weaker spatial representations. Following~\citet{nveagle}, embeddings generated by each encoder were interpolated and concatenated along the channel dimension before resampling. Combining encoders consistently outperforms their single-encoder counterparts, where InternVideo$2+$SigLIP-SO$400$M was the best overall, exhibiting a $\sim7\%$ improvement in \benchmark. We found that video encoders with fewer input frames perform more favorably, possibly due to better image-video transfer. This design is in line with~\citet{qwen2vl}, whose vision encoder encodes videos two input frames at a time.

\vspace{0.05in}
\finding{5}{Combining SigLIP-SO$400$M with InternVideo2 leads to the best overall performance.}

\subsection{Video token resampling}
\label{sec:arch:resamplers}

\begin{table}[b]
    \centering
    \adjustbox{width=\linewidth}{
    \begin{NiceTabular}{lcccccc} 
    \CodeBefore
    \rectanglecolor{metabg}{2-7}{-7}
    \Body
    \toprule
    & \multicolumn{6}{c}{\benchmark} \\
    \cmidrule(lr){2-7}
    \textbf{Connector} & OCR & Spatial &  Egocentric & Perception & Reasoning & Overall \\
    \midrule
    2-layer 2D Conv. + adaptive average pooling~   & 43.0  & 50.5 & 44.5 & 44.0 & 42.0 & 44.7 \\[0.5ex]
    2-layer MLP + adaptive average pooling     & 47.5  & 53.7 & 51.5 & 52.0 & 61.5 & \underline{53.2} \\[0.5ex]
    Perceiver Resampler    & 50.4 & 54.8 & 58.5 & 58.8 & 55.4 & \textbf{55.5} \\
    \bottomrule
    \end{NiceTabular}
    }
    \caption{\textbf{Video token resampling methods.} Performance of different token resampling techniques on video-LMM tasks. The Perceiver Resampler outperforms other methods across all metrics. Different encoder features are concatenated along the channel dimension, following~\citet{nveagle,cambrian1}.}
    \label{table:resamplers}
\end{table}

Vision encoders output vision embeddings in a lower dimensionality than LLMs' hidden dimension, requiring a $2-4\times$ up-projection. Early methods often up-projected all visual tokens directly into the LLM's space~\citep{video_llava,video_chat}. This approach leads to informational waste by instilling a synthetic information bottleneck.
~\citet{idefics2} demonstrated that resampling image tokens (where multiple up-projected tokens are pooled into one) does not reduce performance in image-LMMs. Token resampling is even more critical in video-LMMs as this directly affects how many frames can be processed, limiting the maximum video length.
Video token resampling can be text-guided (e.g., using a Q-Former)~\citep{llama-vid,mvbench,video_llama}. However, this approach does not generalize well to multi-turn conversations, as tokens will be down-sampled according to the first question. Many others do some form of average pooling~\citep{chat_univi,vidcompress,xu2024slowfast}.

\citet{nveagle} tested multiple encoder integration approaches and found that channel-wise concatenation was preferable in nearly all configurations. Therefore, we adopted channel-wise concatenation in our experiments. We tested three token resampling methods: mlp up-projection + average pooling, 2D conv + average pooling, and perceiver resampling. 
As shown in Tab.~\ref{table:resamplers}, the Perceiver Resampler outperforms the other methods across all metrics. While~\citet{idefics3} reported that utilizing the Perceiver Resampler hurts OCR performance; this trend was not observed in videos with the limited available token count per frame.
Another key difference is the initial channel-wise concatenation of encoder features before resampling. This alignment enables the Perceiver to integrate features from different encoders better as they are better spatially aligned.

\vspace{0.05in}
\finding{6}{Perceiver resampling shows superior performance when reducing the tokens/frame.}
\vspace{0.05in}

Some methods utilize active token pooling, where the initial question is used to guide the token pooling~\citep{llama-vid,mvbench,video_llama}, usually using a Q-Former, and were not included in this study. Note that these would require token resampling at every conversational turn.

 \begin{table*}[t]
    \setlength\tabcolsep{2mm}
    \centering
    \adjustbox{width=\textwidth}{
    \begin{NiceTabular}{lcccccc}
    \CodeBefore
    \rectanglecolor{metabg}{2-7}{-7}
    \Body
    \toprule
    & \multicolumn{5}{c}{\benchmark} \\
    \cmidrule(lr){2-7}
    \textbf{Format} & OCR & Spatial & Egocentric & Perception & Reasoning & Overall \\
    \midrule
    \texttt{<vid\_token>} &  50.4 & 54.8 & 58.5 & 58.8 & 55.4 & 55.5 \\[0.5ex]
    \texttt{<vid\_start><vid\_token><vid\_end>} & 49.2 & 54.8 & 61.7 & 60.2 & 57.9 & \underline{56.7} \\[0.5ex]
    \texttt{clip from \{MM:SS\}-\{MM:SS\}:<vid\_token>}  & 50.0 & 54.0 & 61.7 & 60.8 & 57.9 &\textbf{ 56.8} \\[0.5ex]
   \texttt{\shortstack[l]{clip from \{MM:SS\}-\{MM:SS\}: \\ <vid\_start><vid\_token><vid\_end>}} & 
    50.0 & 
    54.2 & 
    61.2 & 
    55.7 & 
    60.6 & 
    56.2 \\
    \bottomrule
    \end{NiceTabular}
    }\vspace{-0.1in}
    \caption{\textbf{Video token integration methods.} Performance of different strategies for integrating video tokens into the text sequence. Incorporating textual timestamps before each clip yields the best overall performance.\label{table:temp_embed}\vspace{-0.15in}
    }
\end{table*}

\subsection{Video token integration}
\label{sec:arch:integration}

Integrating video and text tokens is a pivotal design choice for video-LMMs, as it directly influences how effectively the model processes and interprets multimodal content. Initial works naively concatenated the text and video tokens~\citep{chat_univi,video_chat,video_llava}. However, recent trends have begun to either use separation tokens~\citep{kangaroo} or via text (where a prompt is inserted between frames, usually indicating either frame ID or timestamp)~\citep{llava}. This design choice was also systematically ablated by~\citet{omchat}.
To identify the most robust integration strategy, we experimented with four different methods, which can be seen in Tab.~\ref{table:temp_embed}. 
We evaluated four integration strategies: direct insertion, separation tokens, textual timestamps, and combining separation tokens with timestamps.
As can be seen, we found that adding any text or learnable tokens between video tokens results in a $2-3\%$ improvement across \benchmark. As such, we use the clip timestamps as they do not require learning any new token embeddings. 

\vspace{0.05in}
\finding{7}{Adding tokens (text, learned, etc.) between the video tokens derived from different frames or clips is sufficient for efficient token integration.}

\section{How should video-LMMs be trained?}
\label{sec:training}
This section explores different training schedules and protocols for video Large Multimodal Models (video-LMMs). We begin by testing different training schedules and comparing single to multi-stage training (Sec.~\ref{sec:training:stages}). We then examine when video encoders should be trained and with what data (Sec.~\ref{sec:training:encoders}). Finally, we explore how data composition affects performance (Sec.~\ref{sec:training:data}).

\subsection{Training schedules}\label{sec:training:stages}

We systematically evaluated the impact of different training schedules on model performance, comparing single-stage, two-stage, and three-stage training protocols.
Some studies have suggested that a single-stage training protocol performs similarly to two-stage ones but is more computationally efficient~\citep{prismatic}. However,~\citet{cambrian1} demonstrated that two-stage training improved model performance. Since then, many methods have broken down training into more and more training stages. For example,~\citet{llava,kangaroo} utilized four training stages. 

Video-LMMs are typically trained on a mixture of text, image, multi-image, and video data; it is possible to break down training into even more training steps, each with different components unfrozen and trained on different data compositions. For example, many video-LMMs include an additional, final training stage on long videos as these datasets are expensive and relatively small~\citep{llama-vid,kangaroo}. Others first train on exclusively image datasets before training on multiple modality mixtures~\citep{llava}. We tested seven possible training configurations to evaluate the effect of these different training strategies. 

As shown in Tab.~\ref{table:training_stages}, we found that gradually training the model yields the best performance. Specifically, we found that training the model
over three stages yields the best performance, closely followed by the two-stage training schedules. Please note that different stages have different data compositions; specifically, whenever the LLM is frozen, the other components are tuned only on video data, and when the LLM is tuned, a mixture of text, image, multi-image, and video (following Sec.~\ref{sec:training:data}) is used.

\begin{table*}[t]
    \centering
    \setlength\tabcolsep{1.9mm}
    \adjustbox{width=\textwidth}{
    \begin{NiceTabular}{lccccccccc}
    \CodeBefore
    \rectanglecolor{metabg}{2-10}{10-10}
    \Body
    \toprule
    & \multicolumn{3}{c}{Training Stages} & \multicolumn{6}{c}{\benchmark} \\
    \cmidrule(lr){2-4} \cmidrule(lr){5-10}
    & \lmmicon &  \lmmicon &  \lmmicon  & OCR & Spatial& Egocentric & Perception & Reasoning & Overall \\
    \midrule
    \multirow{2}{*}{1 stage} 
         & \freeze\train\train & - & -  & 42.0 & 46.5& 54.9 & 50.0 & 49.5 & 48.7  \\
         & \train\train\train  & - & - & 28.8 & 29.2 & 18.8 & 35.5 & 22.6 & 30.8 \\
        \midrule 
     \multirow{3}{*}{2 stage} 
         & \freeze\train\freeze & \freeze\train\train & -  & 52.2 & 54.5& 55.9 & 60.3 & 58.4 & 56.3 \\ 
         & \train\train\freeze  & \freeze\train\train & -  & 51.6 & 54.5 & 58.0& 62.1 & 60.2 & \underline{57.8} \\
         & \train\train\freeze  & \train\train\train & - & 42.2 & 48.9 & 61.7 & 43.7 & 52.2 & 48.1 \\ 
    \midrule 
    \multirow{2}{*}{3 stage} 
         & \freeze\train\freeze  & \train\train\freeze & \freeze\train\train   & 53.0 & 52.5 & 64.9 & 59.8 & 65.9 & \textbf{59.2} \\

         & \freeze\train\freeze  & \train\train\freeze & \train\train\train   & 44.2 & 37.5 & 43.9 & 56.6 & 38.5 & 44.2  \\
    \bottomrule
    \end{NiceTabular}
    }\vspace{-0.05in}
    \caption{
     \textbf{Training schedules.} An overview of the seven different training schedules evaluated, highlighting whether the LLM and vision encoders are frozen or unfrozen during each stage and the types of data used for training.
    \freeze $~$ and \train $~$ indicate whether a module is frozen or trainable, respectively.
    For each training schedule, three hyperparameters were tested, and we report the best-performing model.
    \label{table:training_stages}
    }
\end{table*}

\vspace{0.05in}
\finding{8}{Progressively unfreezing the different components in different stages leads to superior model training dynamics.}

\subsection{Training video encoders}\label{sec:training:encoders}
It is unclear when and with what data one should train the video encoders.~\citet{cambrian1} reported that training image encoders is beneficial in image-LMMs. 
However, video-LMMs are trained on a mixture of video, multi-image, and image data. Furthermore, to have a unified encoding scheme, images are replicated $N$ times to be encoded by the video encoder. As such, these models have additional dimensions of the data mixture on which the encoders can be trained. We compared training vision encoders on either the data mixtures or exclusively on video data and whether first aligning the connector improves performance in Tab.~\ref{table:training_stages}.

\begin{figure}[t]
\centering
\includegraphics[width=0.75\linewidth]{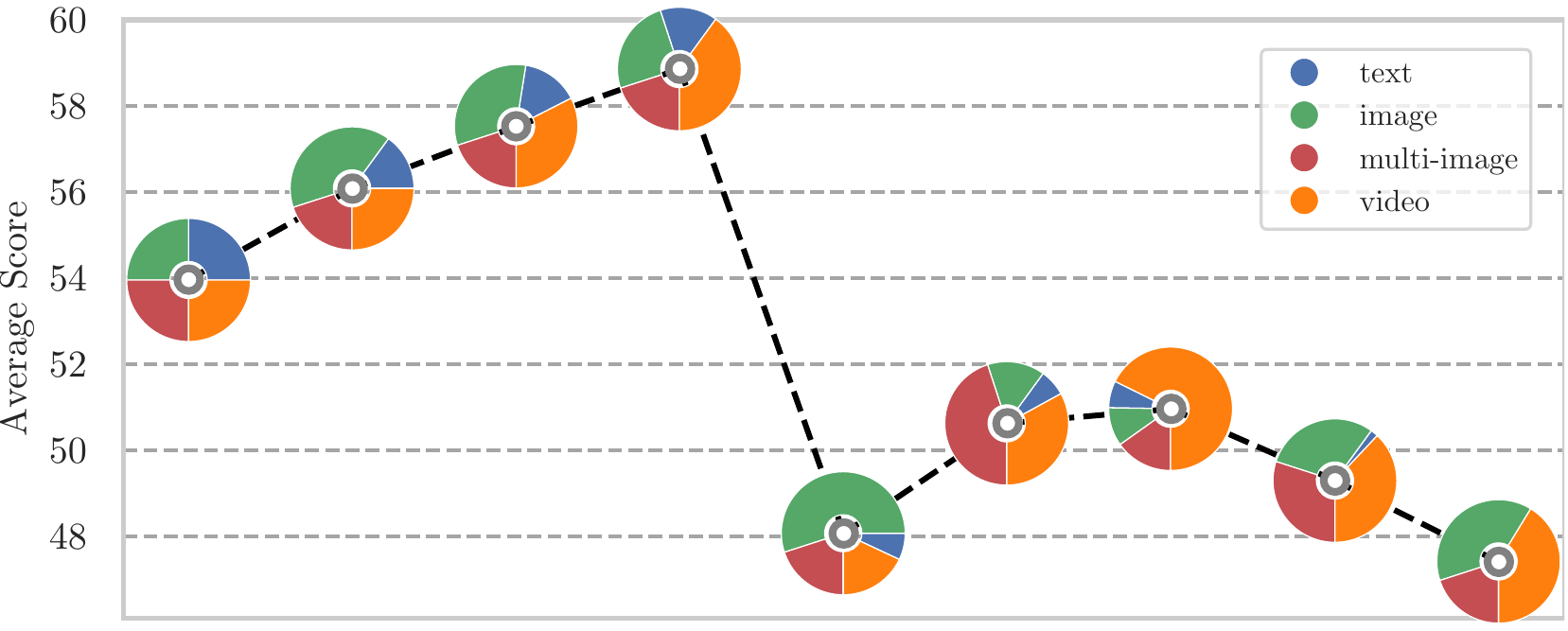}\vspace{-0.07in}
\caption{\label{fig:data_mixture} \textbf{The effect of data mixture on performance.} We find that having $\sim10-14\%$ text data is important for video understanding performance. Out of our $14\%$ data mixtures, we tested image-heavy, balanced, and video-heavy mixtures and found that video-heavy mixtures performed the best.\vspace{-0.07in}
}
\end{figure}

In all experiments, if the LLM is frozen, the model is trained only with video data. When the LLM is unfrozen, we use a data mixture of text, image, multi-image, and video data as described in Sec.~\ref{sec:training:data}. As such, if both the video and LLM are unfrozen simultaneously, the vision encoders will be trained on a combination of image and video data. We found that this significantly hurts LMM performance. Training the encoders improves egocentric reasoning performance while the rest of the metrics remain largely unaffected, most likely due to better fine-grained vision-language alignment.
These insights are in line with~\citet{zhao2024distilling}'s report. 

\vspace{0.05in}
\finding{9}{Finetuning video encoders on only video data further improves overall performance, especially on reasoning and domain-specific tasks.}

\subsection{Data composition}
\label{sec:training:data}

The composition of the training data plays a significant role in the performance of LMMs, as illustrated by~\citet{mm15}.
We investigated how the text, image, and video data mixtures affected video-LMMs performance. Specifically,  we randomly selected several data compositions, as illustrated in Fig.~\ref{fig:data_mixture}. As can be seen, including  $10\sim14\%$ text data in the training mix is required for performance. This likely alleviates catastrophic forgetting. 
Increasing the proportion of text data beyond $14\%$ to $25\%$, or decreasing it below $7\%$, harmed performance.
Beyond including text data, having a slightly video-heavy mix of the remaining modalities was preferable. 
This balance allows the model to learn from higher-quality and diverse image datasets~\citep{llava,video_llava}.

\vspace{0.05in}
\finding{10}{Data mixture matters, and including a moderate amount of text data and maintaining a slight video-heavy mix 
leads to optimal performance.}

\begin{figure*}[t]
\centering
\includegraphics[width=\linewidth]{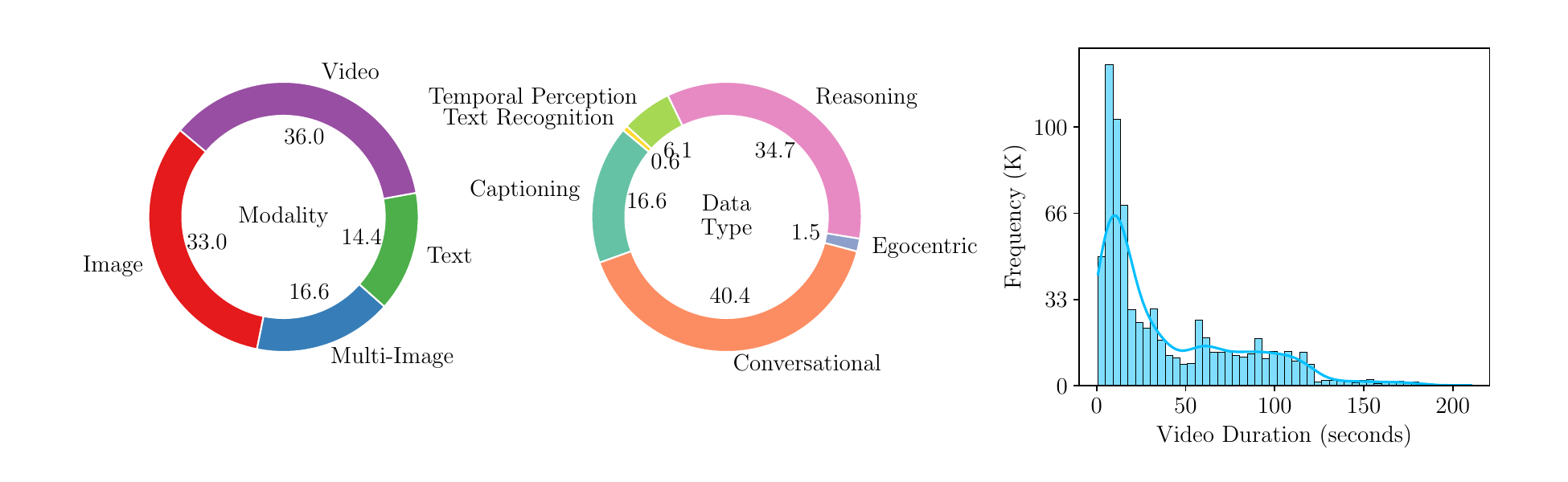}
\caption{
\textbf{Data statistics of the fine-tuning dataset.} 
\textbf{(Left)} Breakdown of data modalities, including text, image, multi-image, and video, illustrating the composition of the fine-tuning dataset. 
\textbf{(Middle)} Distribution of video annotation types, highlighting the proportions of Conversational, Reasoning, Egocentric, Temporal Perception, OCR, and Captioning annotations. 
\textbf{(Right)} Histogram of video durations, showing the distribution of durations in the training dataset.\label{fig:data_comp}
}
\end{figure*}

\section{\method: a family of state-of-the-art large multimodal models}
\label{sec:apollo}
We leverage the findings from our studies and train a family of video-centric Large Multimodal Models (LMMs), \method. \method models have state-of-the-art performance across multiple model sizes, frequently outperforming models $2-3\times$ their size. 
We employed the Qwen$2.5$~\citep{yang2024qwen2} series of Large Language Models (LLMs) at varying scales to serve as the backbone for \method. Specifically, we utilized models with $1.5$B, $3$B, and $7$B parameters. 
Following our analysis in Sec.~\ref{sec:arch}, we used a SigLIP-SO$400$M~\citep{siglip} encoder combined with an InternVideo$2$~\citep{internvideo2} video encoder. Features from each encoder are interpolated and concatenated along the channel dimension before being resampled to $32$ tokens/frame using a Perciver Resampler~\citep{perceiver}. 
We utilized the 3-stage training schedule discussed in Sec.~\ref{sec:training:stages}.

\begin{table*}[t!]
    \centering
\setlength\tabcolsep{2mm}
    \vspace{2mm}
    \resizebox{\textwidth}{!}{%
    \begin{NiceTabular}{l|ccccc|cccccc}
    \CodeBefore
    \rectanglecolor{metabg}{2-12}{37-12}
    \rectanglecolor{metablue!10}{23-1}{23-12}
    \rectanglecolor{metablue!10}{29-1}{29-12}
    \rectanglecolor{metablue!10}{37-1}{37-12}
    \Body
    \toprule
         & \multicolumn{5}{c|}{\textbf{Existing Benchmarks}} & \multicolumn{6}{c}{\textbf{\benchmark}} \\ \\
        \multirow{2}{*}{\textbf{Model}} & 
        \rotatebox{90}{\textbf{\footnotesize{TempCompass}}} & 
        \rotatebox{90}{\textbf{\footnotesize{MLVU}}} & 
        \rotatebox{90}{\textbf{\footnotesize{PerceptionTest}}} & 
        \rotatebox{90}{\textbf{\footnotesize{VideoMME}}} & 
        \rotatebox{90}{\textbf{\footnotesize{L-VideoBench}}} & 
        \rotatebox{90}{\textbf{\footnotesize{OCR}}} & 
        \rotatebox{90}{\textbf{\footnotesize{Egocentric}}} & 
        \rotatebox{90}{\textbf{\footnotesize{Spatial}}} & 
        \rotatebox{90}{\textbf{\footnotesize{Perception}}} & 
        \rotatebox{90}{\textbf{\footnotesize{Reasoning}}} & 
        \rotatebox{90}{\textbf{\footnotesize{Overall}}} \\
        \cmidrule(lr){2-12}
    & mc & m-avg & val & wo/w sub. & val &  &  &  & &  \\ \midrule
    \textit{Proprietary} \\
        GPT-4V~\citep{openai2023gpt4v} & - & 49.2& -    & 59.9/63.3 & 61.3 & 65.7 & 55.0 & 70.8 & 41.0 & 44.7 & 58.7 \\
        GPT-4o~\citep{openai2024gpt4o} & 70.9    & 64.6 & -    & 71.9/77.2 & 66.7 & 76.0    & 69.2   & 90.1   & 82.0   & 83.1    & 79.8 \\
        Gemini-1.5-Flash~\citep{team2023gemini} & -       & -    & -    & 70.3/75.0 & 61.6 & -    & -    & -    & -    & -    & - \\
        Gemini-1.5-Pro~\citep{team2023gemini} & 69.3       & -    & -    & 75.0/81.3 & 64.0 & 74.5    & 77.1   & 79.5    & 85.1    & 88.1    & 80.6 \\
        Claude-3.5-Sonnet~\citep{anthropic2024claude35} & -       & 36.5   & -    & 60.0/62.9 & - & - & - & - & -  & - & - \\
        \midrule
        \textit{Open-weight} \\
        Qwen2VL-2B~\citep{qwen2vl}   & 60.6   & 59.5     & 53.9    &  55.6/60.4           & 48.5 & 29.0 & 29.0 & 47.0 & 50.0 & 46.0 & 40.2 \\    
        Qwen2VL-7B~\citep{qwen2vl}   & 68.5     & 65.5   & 62.3   & 63.3/69.0          & 55.6  & 
        57.4 & 67.5 & 63.7 & 71.2 & 67.9 & 66.0 \\
        Qwen2VL-72B~\citep{qwen2vl}   & -  & -     & 68.0    &  71.2/77.8           & - & - & - & - & -  & - & - \\    
        Aria 8x3.5B~\citep{aria}  & 69.9   & -     & 53.9    &  67.6/72.1           & 64.2 & - & - & - & -  & - & - \\    
        Pixtral-12B~\citep{pixtral}   &-   & -    & -   & 40.7/47.5  & 44.9 & - & - & - & -  & - & - \\    
        \midrule\midrule
        \textit{Open-source} \\

        LLaVA-OV-0.5B~\citep{llava} &  53.2  & 50.3 & 49.2 & 44.0/43.5 & 45.8 & 38.0 & 27.0 & 28.0 & 20.0 & 38.0 & 30.0 \\
        VILA1.5 3B~\citep{vila}   & 56.1     & 44.4    & 49.1    & 42.2/44.2          & 42.9    & 31.7 & 33.0 & 29.3 & 38.0 & 44.7 & 36.1 \\
        InternVL2-2B~\citep{llava} &53.4&48.2&49.6&30.8/-&44.8&40.8 & 46.3 & 34.3 & 44.7 & 45.3 & 42.1 \\
        Phi-3.5-Vision-4.2B~\citep{phi3.5v}   & -     & - & - & 50.8/-    & - &   -  & -& - &-  & -   & - \\
        LongVU 3.2B~\citep{longvu}  & -     & \underline{55.9} & -   & \underline{51.5}/-           & - & - & - & - & - & - & - \\

        \textbf{\method-1.5B}    & \textbf{60.8}     & \textbf{63.3} & \textbf{61.0} & \textbf{53.0/54.6}    & \textbf{54.1} & \textbf{49.0}    & \textbf{63.3} & \textbf{50.0} & \textbf{66.5} & \textbf{57.4}   & \textbf{57.0} \\

        \midrule
        LongVA-7B~\citep{longva} & -       & 56.3 & -    & 52.6/54.3 & -    &32.4 & 43.1 & 41.0 & 37.7 & 51.1 & 41.5 \\
        XComposer-8B~\citep{zhang2024internlm} & -       & 37.3 & 34.4 & 55.8/58.8 & -    &  \textbf{50.7} & 42.0 & 54.7 & 54.7 & 40.5 & 48.6 \\
        Kangaroo-8B~\citep{kangaroo} & 61.3    & 61.0    & -    & \underline{56.0}/57.6         & 54.2   & - & - & - & -  & - & - \\
        Video-XL 7B~\citep{videoxl} & -       & 64.9    & -    & 55.5/\textbf{61.0}      & 49.5    & - & - & - & -  & - & - \\
        Oryx 7B~\citep{oryx} & -        & \underline{67.5}    & -    & 50.3/55.3      & \textbf{55.5}    & - & - & - & -  & - & - \\        
        
        \textbf{\method-3B}    & \textbf{62.5}     & \textbf{68.7} & \textbf{65.0 }& \textbf{58.4}/\underline{60.6}    & \underline{55.1} & \underline{49.6}   & \textbf{68.6}   & \textbf{59.3}    & \textbf{67.0}   & \textbf{68.4}    & \textbf{62.7} \\
        \midrule
        InternVL2-8B ~\citep{internvl2}   & 65.3     & 50.8    & 57.4   & 54.0/56.9          & 51.8    &  50.0 & 48.4 & 54.3 & 57.7 & 51.8 & 52.8  \\ 
        LLaVA-OV-7B~\citep{llava}  & 64.8      & 64.7 & 57.1 & 58.2/61.5 & 56.4 & \textbf{56.0} & \textbf{69.1} & \textbf{69.0} & \underline{63.3} & \underline{63.2} & \underline{64.0} \\
        LongVU 7B~\citep{longvu} & -       & 65.4    & -    & 60.6/-      & -    & - & - & - & -  & - & - \\
        LLaVA-N-Video-32B~\citep{zhang2024llavanextvideo} & -    & 39.3    & 59.4 & 60.2/63.0 & 50.5 & -    & -    & -    & -    & -    & - \\

        Oryx 34B~\citep{oryx} & -     & \underline{70.8}   & -    & 53.9/58.0      & \textbf{62.2}   & - & - & - & -  & - & - \\
        VILA-1.5-40B~\citep{vila} & -      & 56.7    & 54.0 & 60.1/61.1 & -    &  - & - & - & -  & - & -  \\

        InternVL2-34B~\citep{internvl2} & -     & 59.9    &-   & \underline{61.2}/62.4          & -    & -    & -    & -    & -    & -    & - \\
        \textbf{\method-7B}   & \textbf{64.9}       & \textbf{70.9} & \textbf{67.3} & \textbf{61.3/63.3}    & \underline{58.5} & \underline{51.6}   & \underline{68.4}    & \underline{67.5}    & \textbf{69.8}    & \textbf{71.2}    & \textbf{66.3} \\
        \bottomrule
    \end{NiceTabular}%
    }
        \caption{\textbf{Performance of \method on a diverse range of video benchmarks.} 
We compare~\method to both proprietary and open-source models across multiple benchmark suites and our curated~\benchmark. 
\textbf{(Top)}~\method-$1.5$B surpasses various small LMMs, including those with larger parameter counts (e.g., LongVU $3.2$B), demonstrating robust gains even at relatively modest scale. 
\textbf{(Middle)}~\method-$3$B maintains impressive results and competes effectively with recent $7$B models such as Oryx and Video-XL, underscoring the efficiency of our design decisions in bridging performance gaps without massive scaling. 
\textbf{(Bottom)}~\method-$7$B attains state-of-the-art performance among models of a similar size and even outperforms some models with over $30$B parameters, highlighting its robustness.~\label{tab:results:small_models}}
\end{table*}

We curated a diverse mixture of publicly available and licensed datasets spanning text, image-text, multi-image, and video modalities. Due to licensing constraints, we omitted non-permissive sources (e.g., those reliant on ChatGPT), limiting the inclusion of some commonly used datasets. To further enhance our training corpus, we generated multi-turn video-based conversations via an annotation tool powered by LLaMA $3.1$ $70$B~\citep{touvron2023llama}. Figure~\ref{fig:data_comp} provides a detailed overview of our data composition and statistics.

We evaluated \method across a suite of benchmarks to assess its performance in video-language understanding tasks, including TempCompass~\citep{tempcompass}, MLVU~\citep{mlvu}, Perception-Test~\citep{perceptiontest}, Video-MME~\citep{videomme}, LongVideoBench~\citep{longvideobench}, and \benchmark. 
As shown in Tables~\ref{tab:results:small_models}, \method models demonstrate strong performance across benchmarks. Notably, \method-$3$B outperforms several recently introduced $7$B models, such as Oryx-$7$B~\citep{oryx}, Kangaroo~\citep{kangaroo}, and Video-XL-$7$B~\citep{videoxl}. For instance, on the MLVU benchmark, \method-$3$B achieves a score of $68.7$, surpassing Oryx-$7$B's $67.5$. Similarly, \method-$1.5$B outperforms models larger than itself, including Phi-$3.5$-Vision ($4.2$B parameters) and some $7$B models like LongVA-$7$B~\citep{longva}, indicating that smaller models can suffice for proof-of-concept implementations. We hope these results will motivate the field to utilize such smaller models for faster prototyping in the future.

Furthermore, \method-$7$B establishes a new performance frontier for models at the $7$B scale, rivaling and even surpassing models with over $30$B parameters such as Oryx-$34$B and VILA$1.5$-$40$B~\citep{vila}. On the MLVU benchmark, for instance, \method-$7$B scores $70.9$, narrowly outperforming Oryx-$34$B's $70.8$. These gains highlight the potency of our design insights and confirm that carefully chosen architectural and training strategies can yield substantial improvements without resorting to larger model sizes.

\section{Background}
\paragraph{Video Large Multimodal Models.}
Early video-LMMs~\citep{frozenbilm, zhu2023minigpt,maaz2023video, xu2024pllava} relied on sparsely sampled frames and MLP connectors or entirely training-free methods~\citep{image_grid,freeva}. To address token count and support long-form video understanding, subsequent works introduced resampling methods such as spatio-temporal pooling~\citep{video_llama, longvu, xu2024pllava, chat_univi, llava_video, xu2024slowfast}.
Most approaches~\citep{ccam,chat_univi,kangaroo,oryx,longvu} use image-based encoders, with only a few~\citep{video_llava,internvl2,mvbench} employing video-specific encoders to capture temporal dependencies. Training schedules typically involve alignment followed by supervised fine-tuning~\citep{video_llava,llava,internlmxcomposer2_5,videoxl} to adapt connectors and LLMs for video understanding.
Earlier video-LMMs~\citep{video_llava,video_llama} were trained on small-scale video instruction datasets, while recent efforts have expanded both dataset scale~\citep{llava_video,oryx} and quality~\citep{sharegpt4video}, leveraging multi-image datasets~\citep{vila,llava,videoxl} to enhance model capabilities further. 
Benchmarks have also evolved, shifting from short-video tasks~\citep{activitynetqa,msvdqa,msrvttqa} to long-video tasks~\citep{mlvu,longvideobench,temporalbench}.
Despite these advances, many design decisions in video-LMM remain with limited analysis or justification. This work addresses these gaps by systematically exploring the design space for video-LMMs.

\paragraph{Design Exploration for Large Multimodal Models.}
Recent studies have highlighted the importance of systematically exploring the design space for image-based LMMs~\citep{prismatic,idefics2,cambrian1,nveagle}, focusing on key components such as encoder selection, training strategies, and data mixtures. \citet{idefics2} introduced perceiver resamplers as an effective method for reducing token counts and enabling efficient long-context modeling. ~\citet{prismatic, cambrian1} examined trade-offs between single versus multiple encoders, training versus freezing encoders, and one-stage versus multi-stage training, with Cambrian-1 also analyzing the influence of data mixtures and vision-centric benchmarking.~\citet{nveagle} extended these efforts by evaluating the impact of diverse encoder architectures and their combinations.
While these works provide a strong foundation for image-based LMMs, the design space for video-LMMs remains underexplored. Unlike images, videos require specialized strategies for frame sampling, token resampling, encoder selection, and efficient training and evaluation. This work addresses these gaps by systematically investigating the unique challenges and opportunities in designing video-LMMs, paving the way for scalable and effective solutions in video understanding.

\section{Conclusion}

In the study, we critically evaluated the current state of the video Large Multimodal Model (video-LMM) field, from architecture design and training schedules to data mixtures and evaluation. In part, we hope that concepts such as \designscale encourage researchers to utilize smaller LMMs in their research, while \benchmark will allow for faster and more comprehensive evaluation. We hope our insights into the key aspects of video-LMM design, encompassing video sampling, encoder selection, token resampling, and token integration, will further democratize video-LMM research, further accelerating research in the field. 

Building upon these insights, we developed \method, a family of state-of-the-art LMMs capable of advanced video-language understanding. Notably, \method-$3$B outperforms most advanced $7$B models, while \method-$7$B outperforms all $7$B models and many recent $30$B models. Our findings highlight that careful design and training strategies can yield superior performance without necessitating larger model sizes. We believe that our work provides valuable guidelines and resources for future research, advancing the development of efficient and effective video-LMMs.

\newpage

\bibliographystyle{assets/plainnat}
\bibliography{paper}

\clearpage
\newpage
\beginappendix

This document provides more details of our approach and additional experimental results, organized as follows:
\begin{itemize}
	\vspace{2pt}
 	\item \S~\ref{app:sec:benchmark_analysis} \textbf{Analyzing the benchmarks.} We provide an in-depth analysis of the factors affecting evaluations, such as video duration and format. We then give a detailed overview of how we curated \benchmark.
	\item \S~\ref{app:sec:implementation} \textbf{\method implementation details.} We provide an in-depth description of \method, along with all the hyperparameters needed to reproduce \method.
 	\item \S~\ref{app:sec:scaling} \textbf{\designscale.} We provide an in-depth analysis of the correlations between models of different sizes, compare \designscale to traditional scaling laws, and motivate their usage in future experiments.
 	\item \S~\ref{app:sec:sampling} \textbf{Video sampling analysis.} We expand on our Video Sampling experiments and add a per-metric breakdown.
 	\item \S~\ref{app:sec:raw} \textbf{Raw results.} We provide all the raw data used in our study for further analysis. 
    For Sec.~\ref{section:scaling-laws}: Tab.~\ref{tab:results:raw_scaling1} \&~\ref{tab:results:raw_scaling2},
    Sec.~\ref{sec:arch:sampling}: Tab.~\ref{tab:fps_sampling_raw} \&~\ref{tab:uniform_sampling_raw},
    Sec.~\ref{sec:arch:encoders}: Tab~\ref{tab:vision_encoder_raw},
    Sec.~\ref{sec:arch:integration}: Tab.~\ref{table:temp_embed}, 
    Sec.~\ref{sec:training:stages}: Tab.~\ref{table:training_stages_raw}, 
    Sec.~\ref{sec:training:data}: Tab.~\ref{tab:data_mix_raw}.
\end{itemize}

\part{}

{
\hypersetup{linkcolor=black} 
\parttoc 
}

\section{Future work}

Several promising directions emerge from our study on Large Multi-modal Models (LMMs). First, we employed a fully unified architecture, using the video encoder for videos and images by replicating images N times. Exploring separated architectures, where images are processed with an image encoder and videos with both image and video encoders, could reveal performance benefits and better modality handling.

Second, in separated architectures, training the video and image encoders during supervised fine-tuning (SFT) and evaluating their individual contributions to performance could identify optimal training strategies. Similarly, training both encoders on mixed image and video data within unified architectures may help determine which encoder influences observed performance drops, enabling targeted improvements.

Further investigation into \designscale is necessary to confirm its applicability across a broader range of model sizes, ensuring its reliability for even larger models. We did not explore memory-based LMM approaches, such as memory banks or frame retrieval methods like text-conditioned pooling in Q-Former. Evaluating these techniques could test our hypothesis that they may not generalize well to multi-turn conversations.

Lastly, current benchmarks primarily use academic multiple-choice formats, which inadequately assess conversational abilities. Developing a dedicated conversational evaluation benchmark for LMMs is essential to more accurately measure and enhance models’ dialogue performance in real-world scenarios.

\begin{figure}[t]
\centering
\includegraphics[width=\linewidth]{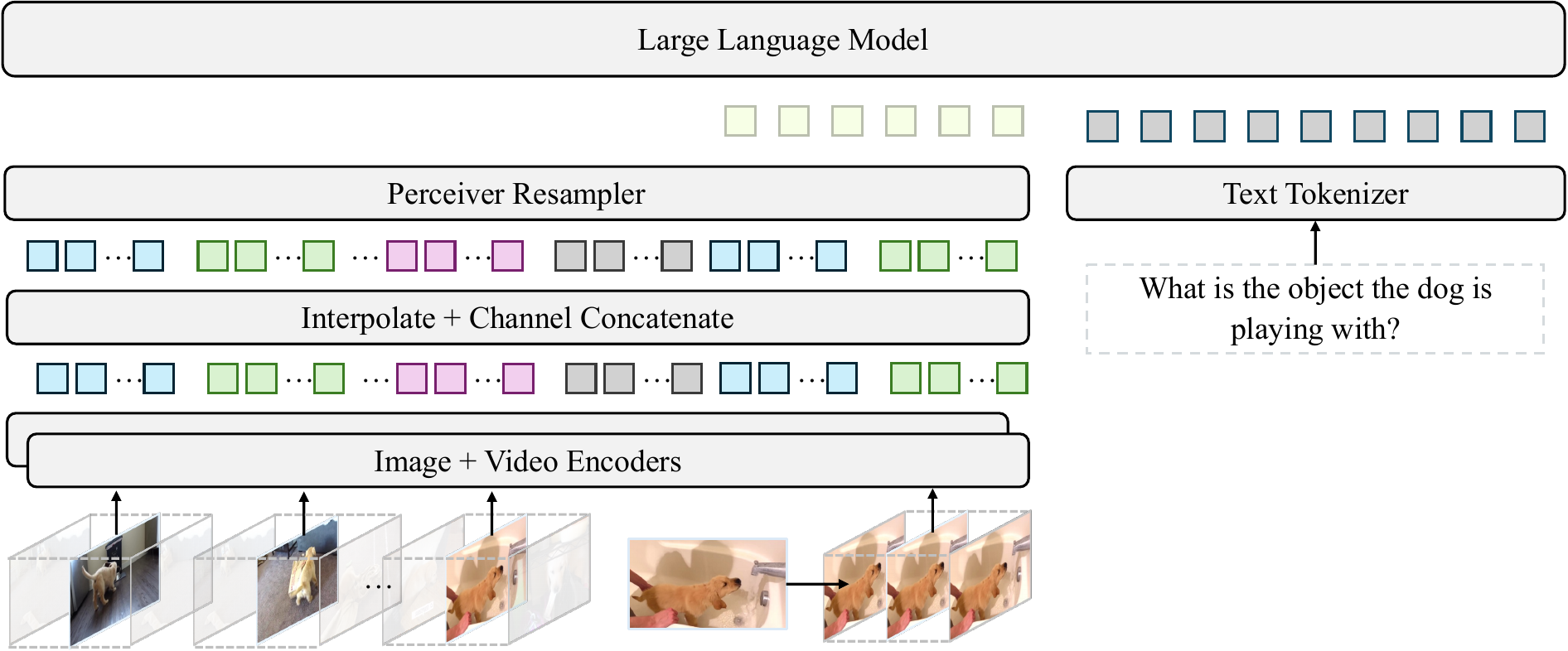}
\caption{\label{fig:apollo_arch} \textbf{\method architecture overview.} \method encodes clips of N  (dependent on the video encoder) frames. Output features are interpolated and concatenated along the channel dimension before being fed to a connector. The connector up-projects the features to the Large Language Models' hidden dimension and then resamples them into a pre-set number of T tokens/clip. Images are duplicated N times and encoded the same way as video clips.  
}
\end{figure}

\begin{figure}[t]
\centering
\includegraphics[width=\linewidth]{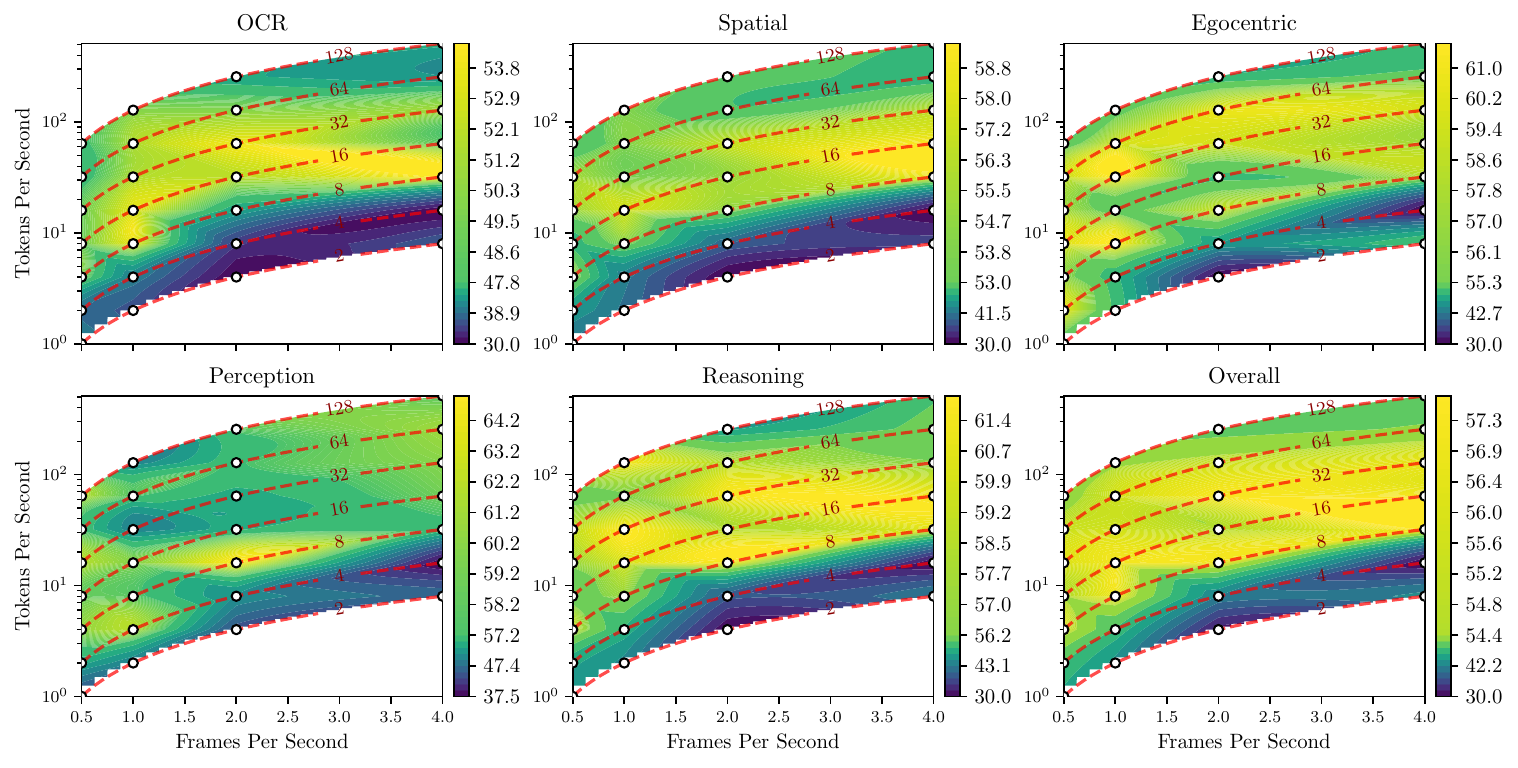}\vspace{-0.1in}
\caption{\label{sup:fig:full_sampling} \textbf{Video fps sampling analysis.} Full analysis on the effect of frames per second (fps, x-axis), tokens per second (tps, y-axis), and tokens per frame (tpf, dotted red lines) on each of \benchmark's dimensions. The number of tokens/frames is highlighted via the dotted red lines.}
\end{figure}

\begin{table}[b]
    \centering
    \adjustbox{width=\textwidth}{
    \begin{tabular}{lcccccccccccccccccccccccccccccc}
    \toprule
    & \multicolumn{3}{c}{NExT-QA} & \multicolumn{3}{c}{Perception-Test} & \multicolumn{3}{c}{TempCompass (CM)} & \multicolumn{3}{c}{TempCompass (MC)} & \multicolumn{3}{c}{TempCompass (YN)} \\
    \cmidrule(lr){2-4} \cmidrule(lr){5-7} \cmidrule(lr){8-10} \cmidrule(lr){11-13} \cmidrule(lr){14-16}
    Model & Video & Image & Text & Video & Image & Text & Video & Image  & Text & Video & Image   & Text & Video & Image & Text\\
\midrule
   InternVL2 2B~\cite{internvl2} &          68.9 &          61.1 &         42.8 &                  49.6 &                  46.0 &                 38.6 &                   67.2 &                   63.3 &                  51.9 &                   53.4 &                   47.5 &                  35.9 &                   62.3 &                   59.3 &                  51.2 \\
  InternVL2 8B~\cite{internvl2} &          70.8 &          72.6 &         49.1 &                  57.4 &                  52.8 &                 41.3 &                   77.4 &                   66.9 &                  58.3 &                   65.3 &                   54.9 &                  43.7 &                   68.6 &                   62.6 &                  52.1 \\
LLaVA-OV 0.5B~\cite{llava} &          57.3 &          50.7 &         31.9 &                  49.1 &                  44.8 &                 40.4 &                   61.9 &                   58.9 &                  51.3 &                   53.2 &                   44.6 &                  34.1 &                   60.0 &                   55.9 &                  49.7 \\
  LLaVA-OV 7B~\cite{llava} &          79.3 &          70.0 &         48.7 &                  57.1 &                  49.7 &                 41.4 &                   73.8 &                   60.8 &                  56.8 &                   64.9 &                   51.6 &                  41.4 &                   69.8 &                   57.8 &                  53.3 \\
     LongVA 7B~\cite{longva} &          50.2 &          38.9 &         36.6 &                  50.6 &                  50.3 &                 50.1 &                   60.7 &                   51.1 &                  50.9 &                   56.1 &                   52.2 &                  50.7 &                   62.9 &                   61.6 &                  60.9 \\
    Qwen2-VL 2B~\cite{qwen2vl} &          68.7 &          62.1 &         44.0 &                  53.1 &                  47.5 &                 39.8 &                   70.9 &                   62.5 &                  54.3 &                   60.6 &                   50.4 &                  40.1 &                   63.7 &                   58.6 &                  52.3 \\
    Qwen2-VL 7B~\cite{qwen2vl} &          78.9 &          68.5 &         42.6 &                  58.9 &                  52.6 &                 38.4 &                   76.6 &                   64.3 &                  56.5 &                   67.2 &                   52.3 &                  41.6 &                   71.9 &                   61.8 &                  54.0 \\
       VILA-1.5 3B~\cite{vila} &          56.9 &          56.7 &        30.1  &                  49.1 &                  49.1 &                36.2  &                   66.3 &                   66.3 &                 52.9  &                   56.1 &                   56.1 &                  36.8 &                   63.4 &                   63.4 &                  51.1 \\
       VILA-1.5 8B~\cite{vila} &          63.1 &          63.1 &        38.2  &                  54.7 &                  54.7 &                41.2  &                   58.7 &                   58.7 &              33.6     &                   49.0 &                   49.0 &                 18.8  &                   62.5 &                   62.5 &                 50.6  \\
  XComposer-8B~\cite{internlmxcomposer2_5} &          71.1 &          47.3 &         41.0 &                  55.9 &                  45.3 &                 39.6 &                   72.2 &                   59.3 &                  49.2 &                   61.1 &                   39.4 &                  31.7 &                   64.5 &                   57.8 &                  52.3 \\
    \bottomrule
    \end{tabular}
    }
    \caption{
    \textbf{Benchmark evaluation for different models across input modalities (1/2).} This table reports the performance of various models on the NExT-QA, Perception-Test, and TempCompass benchmarks with video, image, and text inputs.
    }
    \label{table:group1_benchmark_scores}
\end{table}

\begin{table}[b]
    \centering
    \adjustbox{width=\textwidth}{
    \begin{tabular}{lcccccccccccccccccccccccccccccc}
    \toprule
    & \multicolumn{3}{c}{LongVideoBench} & \multicolumn{3}{c}{MLVU} & \multicolumn{3}{c}{Video-MME (Long)} & \multicolumn{3}{c}{Video-MME (Medium)} & \multicolumn{3}{c}{Video-MME (Short)} \\
    \cmidrule(lr){2-4} \cmidrule(lr){5-7} \cmidrule(lr){8-10} \cmidrule(lr){11-13} \cmidrule(lr){14-16}
    Model & Video & Image & Text & Video & Image & Text & Video & Image  & Text & Video & Image   & Text & Video & Image & Text\\
    \midrule
   InternVL2 2B~\cite{internvl2} &                 44.8 &                 37.9 &                32.8 &       48.2 &       41.5 &      32.6 &                   33.1 &                   30.9 &                  31.4 &                     38.2 &                     32.2 &                    28.7 &                    51.3 &                    39.1 &                   32.8 \\
  InternVL2 8B~\cite{internvl2} &                 51.8 &                 45.0 &                40.2 &       50.8 &       40.0 &      37.5 &                   42.0 &                   40.0 &                  38.6 &                     50.6 &                     39.6 &                    38.6 &                    62.1 &                    48.2 &                   39.4 \\
LLaVA-OV 0.5B~\cite{llava} &                 46.0 &                 40.5 &                37.4 &       50.3 &       39.2 &      35.3 &                   37.2 &                   31.3 &                  33.1 &                     40.0 &                     32.0 &                    30.2 &                    54.6 &                    37.1 &                   30.1 \\
  LLaVA-OV 7B~\cite{llava} &                 56.5 &                 45.1 &                41.2 &       65.1 &       50.3 &      45.5 &                   49.9 &                   36.9 &                  39.8 &                     54.6 &                     39.4 &                    38.3 &                    70.9 &                    47.4 &                   40.2 \\
     LongVA 7B~\cite{longva} &                 45.2 &                 44.2 &                43.0 &       51.9 &       45.1 &      44.1 &                   41.4 &                   38.1 &                  36.7 &                     45.9 &                     39.9 &                    38.4 &                    55.1 &                    45.3 &                   40.0 \\
    Qwen2-VL 2B~\cite{qwen2vl} &                 48.5 &                 40.8 &                40.4 &       59.5 &       45.1 &      38.4 &                   43.2 &                   36.9 &                  33.3 &                     51.0 &                     35.0 &                    32.3 &                    65.3 &                    40.4 &                   34.8 \\
    Qwen2-VL 7B~\cite{qwen2vl} &                 54.8 &                 44.7 &                41.5 &       65.5 &       49.1 &      42.4 &                   49.8 &                   40.0 &                  38.4 &                     57.6 &                     41.2 &                    39.2 &                    70.7 &                    46.3 &                   37.6 \\
   VILA-1.5 3B~\cite{vila} &                 42.9 &                 42.9 &               33.8  &       23.3 &       23.3 &     13.6  &                   31.6 &                   28.0 &                 28.0  &                     36.7 &                     27.3 &                   27.3  &                    48.7 &                    27.8 &                   27.8 \\
       VILA-1.5 8B~\cite{vila} &                 47.2 &                 47.2 &               37.1  &       44.4 &       44.4 &     31.1  &                   39.3 &                   36.6 &                 36.6  &                     42.1 &                     32.3 &                   32.3  &                    56.3 &                    34.3 &                   34.3 \\
  XComposer-8B~\cite{internlmxcomposer2_5} &                 47.6 &                 30.0 &                32.0 &       37.2 &        8.5 &       7.3 &                   46.4 &                   28.0 &                  35.1 &                     50.9 &                     26.3 &                    35.0 &                    66.0 &                    28.1 &                   36.1 \\
    \bottomrule
    \end{tabular}
    }
    \caption{
    \textbf{Benchmark evaluation for different models across input modalities (2/2).} This table reports the performance of various models on the LongVideoBench, MLVU, and Video-MME benchmarks with video, image, and text inputs.
    }
    \label{table:group2_benchmark_scores}
\end{table}

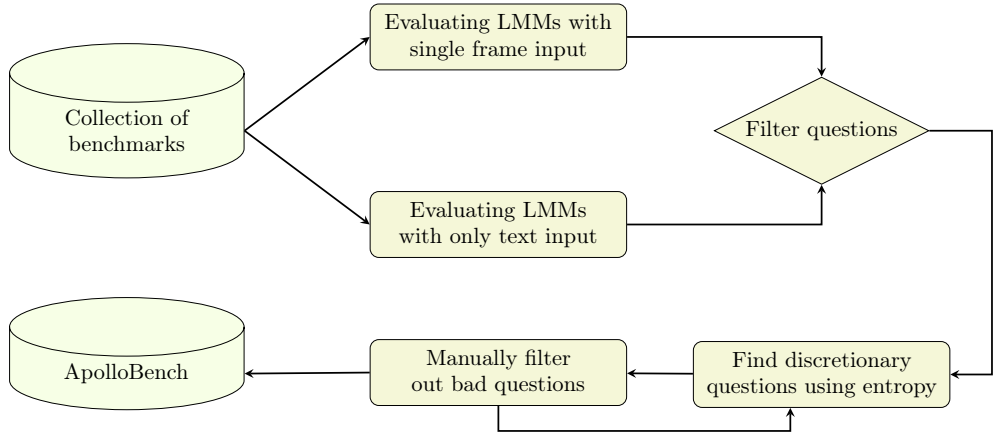
\begin{figure}[t]
    \centering
     \adjustbox{width=0.8\columnwidth}{
\begin{tikzpicture}[node distance=4cm, auto]

    \node (collection) [database] {Collection of benchmarks};

    \node (eval1) [process, right=2cm of collection, yshift=1.5cm, anchor=west] {Evaluating LMMs with single frame input};

    \node (eval2) [process, right=2cm of collection, yshift=-1.5cm, anchor=west] {Evaluating LMMs with only text input};

    \node (filter) [decision, right=7.5cm of collection] {Filter questions};

    \node (entropy) [process, below=2.5cm of filter] {Find discretionary questions using entropy};

    \node (manual) [process, below=1.3cm of eval2] {Manually filter out bad questions};

    \node (apollo) [database, below=1.9cm of collection] {ApolloBench};

    \draw [arrow] (collection.east) -- (eval1.west);
    \draw [arrow] (collection.east) -- (eval2.west);

    \draw [arrow] (eval1.east) -| ([xshift=0cm]filter.north);
    \draw [arrow] (eval2.east) -| ([xshift=0cm]filter.south);
    \draw [arrow] (filter.east) -- ++(1, 0) -- ++(0, -1) |- ([xshift=0cm]entropy.east);
    \draw [arrow] (entropy) -- (manual);
    \draw [arrow] (manual.south) -- ++(0, 0) |- ++(0, -0.4) -| ([xshift=-0.5cm]entropy.south);

    \draw [arrow] (manual.west) -- (apollo.east);
\end{tikzpicture}
}
\caption{\textbf{Flowchart illustrating the curation process of \benchmark.} Starting with a collection of benchmarks, we evaluate Large Multimodal Models (LMMs) using the full video, single-frame, and text inputs. Questions requiring video perception were filtered based on model performance, and discretionary questions were identified using entropy. After manual verification and categorization into five temporal perception categories, the top $400$ questions were selected for the benchmark, and manually inspected.}
    \label{fig:benchmark_creation_flowchart}
\end{figure}

\begin{figure}[t]
\centering
\includegraphics[width=\linewidth]{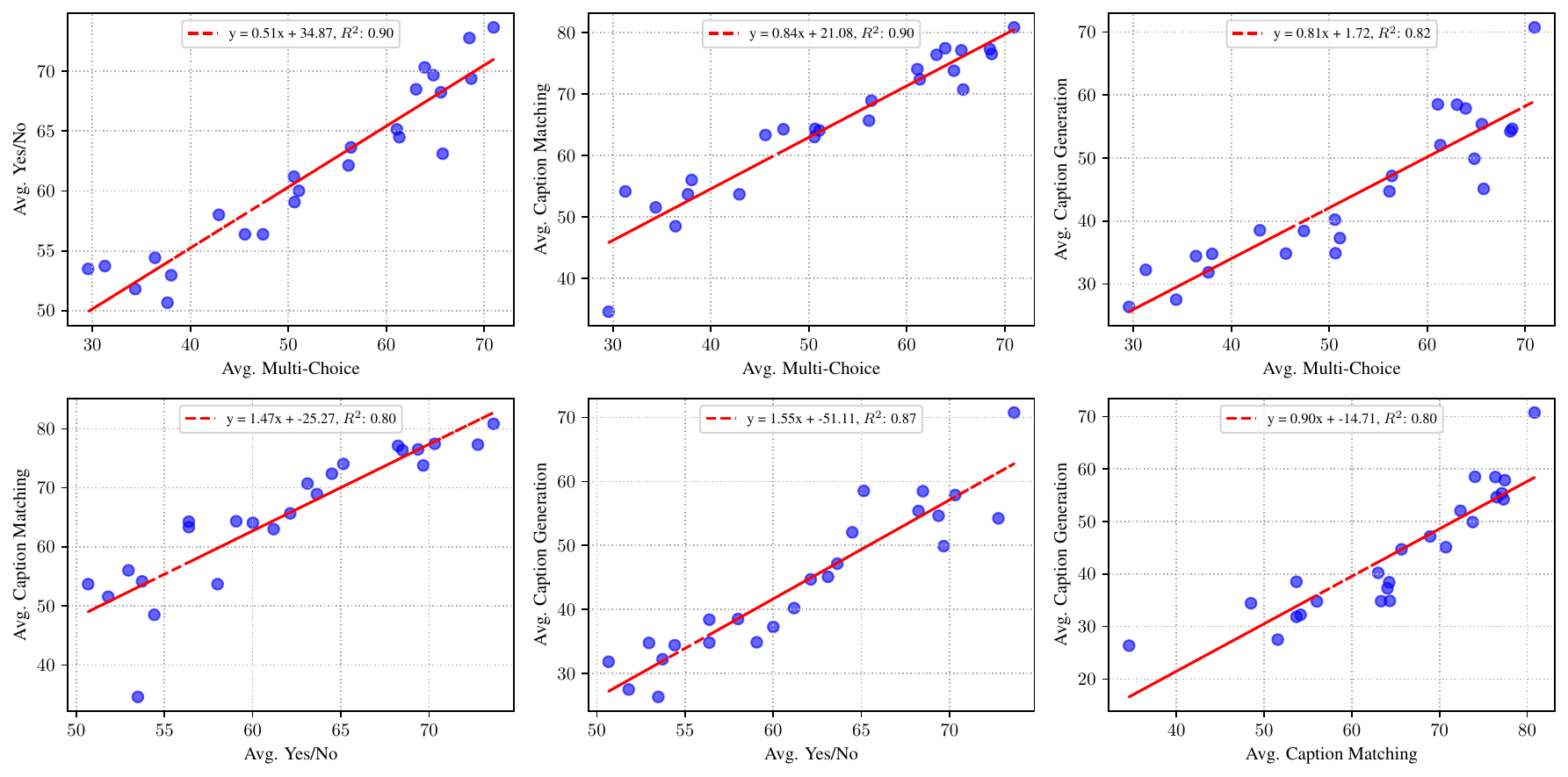}
\caption{\label{sup:fig:eval:tempcompass} \textbf{Effect of question type on model performance.} Correlations between different question types (multiple-choice, yes/no) on the TempCompass are shown. The high correlation indicates consistency in evaluating model performance across various question formats, indicating that multiple choice is a reasonable option in existing benchmarks.}
\end{figure}

\begin{figure}[t]
\centering
\includegraphics[width=\linewidth]{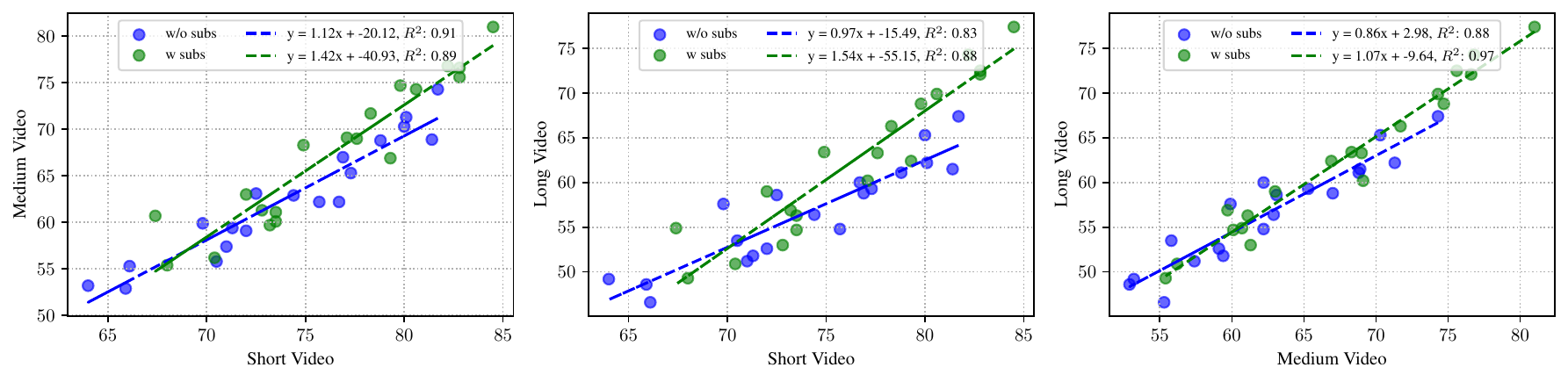}
\caption{\label{sup:fig:eval:videomme} \textbf{Correlation between Video-MME duration groups.} The correlations between short, medium, and long video duration groups on the Video-MME benchmark. The analysis highlights how model performance scales with video length, emphasizing the reliance on text and image modalities as video duration increases.}\end{figure}

\begin{figure*}[t]
\centering
\includegraphics[width=\linewidth]{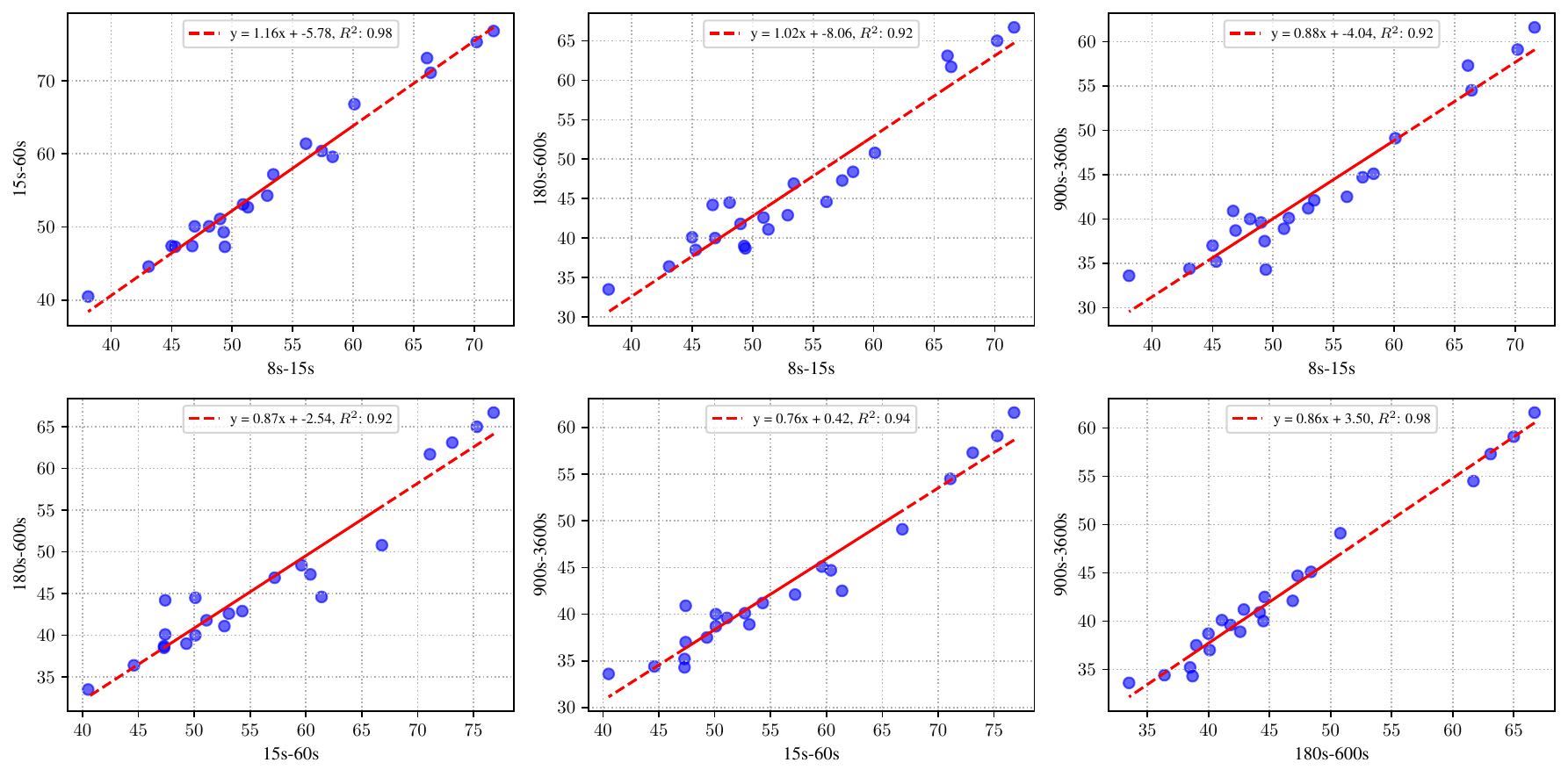}
\caption{\label{sup:fig:eval:longvideobench} \textbf{Correlation between LongVideoBench duration groups.} Correlations between different video duration categories on LongVideoBench are depicted, with $R^2 > 0.92$ across groups. This consistency suggests that performance trends remain stable across varying video lengths.}\end{figure*}

\section{Analyzing the benchmarks}
\label{app:sec:benchmark_analysis}

\subsection{Correlations within existing benchmarks}
\paragraph{Video Duration.}
We were interested in how video length affected model performance to see if existing benchmarks test long video perception capabilities. In the large language model field, testing long-context has been non-trivial, where many benchmarks do not need information integration across the entire model's context window and instead devolve to effectively needle-in-a-haystack style experiments. We hypothesized that long video benchmarks may behave similarly. As such, we compared Video-MME short/medium/long and LongVideoBench's different duration groups (see Fig.~\ref{sup:fig:eval:videomme} and Fig.~\ref{sup:fig:eval:longvideobench}). We found that the two are highly correlated, where $R^2>0.92$ between all duration groups in LongVideoBench (Fig.~\ref{sup:fig:eval:longvideobench}). On Video-MME, whether using or not using subtitles, $R^2>0.83$. When closely examining Video-MME short/medium/long in Fig.~\ref{fig:benchmark_analysis}, one can see that the most significant difference between them is the video modality performance decreasing, with text and image modalities being mostly unchanged. This indicated a greater and greater reliance on the text model's performance rather than any vision capabilities. 

\paragraph{Question types.} There are currently two prevalent methods for evaluating LMMs—either open-ended questions or close-ended (multiple choice, yes/no). Scoring open-ended QA is challenging because the score is ultimately subjective. The dominant way of evaluating open-ended QA is using another language model (e.g., chatGPT) to rate the prediction and decide if it is correct. As shown by~\citet{freeva}, GPT versioning strongly impacts the resulting scores that are even 10\% apart. As such, recent trends show greater reliance on multiple-choice QA. However, are we losing something when evaluating methods only on multiple-choice? As seen in Fig.~\ref{sup:fig:eval:tempcompass}, we find these are highly correlated, with $R^2>0.81$. 
While multiple-choice appears to be a good option for benchmarking the video perception capabilities of video-LMMs, models overly optimized to multiple-choice will not be good conversational agents. As such, a benchmark focusing solely on a conversation is needed, ideally, one that does not suffer from high API costs and GPT versioning noise.

\subsection{Raw evaluations}
We evaluated InvernVL2 2\&8 B~\citep{internvl2}, LLaVA LLaVA-OV 0.5 \& 7B~\citep{llava}, VILA-1.5 1.5 3 \& 8B~\citep{vila}, Qwen2-VL 2 \& 7B~\citep{qwen2vl}, LongVA 7B~\citep{longva} and XComposer-8B~\citep{internlmxcomposer2_5} on NExTQA~\citep{nextqa}, PerceptionTest~\citep{perceptiontest}, TempCompass~\citep{tempcompass}, Video-MME~\citep{videomme}, MLVU~\citep{mlvu}, and LongVideoBench~\citep{longvideobench}. All evaluations were done using lmms-eval~\citep{lmmseval}. Full evaluations of all models on the benchmarks can be seen in Tab.~\ref{table:group1_benchmark_scores} \&~\ref{table:group2_benchmark_scores}.

\subsection{\benchmark curation}

The creation process of \benchmark is depicted in Fig.~\ref{fig:benchmark_creation_flowchart}. The process begins with a collection of multiple-choice benchmarks. To eliminate the reliance on external tools like ChatGPT, we focus exclusively on multiple-choice questions, ensuring a cost-effective and consistent evaluation process~\cite{freeva}.

We first evaluated several Large Multi-modal Models (LMMs) with text-only, center-frame, and full-video inputs. Questions that could be answered correctly by more than $50\%$ of the models using either of these modalities were filtered out, as these questions did not require video perception. 
Next, we categorized the remaining questions into five temporal perception categories: Temporal OCR, Egocentric, Spatial, Perception, and Reasoning. Using entropy, we identified questions with high discrimination power between models and manually verified them to ensure accuracy and quality. From this, we selected the top $400$ questions with the highest entropy to form the final \benchmark dataset. 
This curated benchmark is $41\times$ faster to evaluate compared to existing benchmarks while maintaining a high correlation with their results (see Fig.~\ref{fig:benchmark_analysis}, right). Additionally, \benchmark emphasizes video perception, as shown in Fig.~\ref{fig:benchmark_analysis}, left.

\begin{table*}[htp]
    \centering
    \setlength{\tabcolsep}{6pt}
    \resizebox{\textwidth}{!}{%
    \begin{NiceTabular}{@{}ll|ccc|ccc|ccc@{}}
    \toprule
    & & \multicolumn{3}{c|}{\textbf{Align}} & \multicolumn{3}{c|}{\textbf{Vision Pretraining}} & \multicolumn{3}{c}{\textbf{SFT}} \\ 
    \cmidrule(lr){3-5} \cmidrule(lr){6-8} \cmidrule(lr){9-11}
    & & \textbf{1.5B} & \textbf{3B} & \textbf{7B} & \textbf{1.5B} & \textbf{3B} & \textbf{7B} & \textbf{1.5B} & \textbf{3B} & \textbf{7B} \\
    \midrule 
    \multirow{4}{*}{\rotatebox[origin=c]{90}{\footnotesize \textit{Sampling}}}
    & \textbf{Max clips} & 25 & 25 & 25 & 25 & 25 & 25 & 200 & 200 & 150 \\
    & {fps} & 2 & 2 & 2 & 2 & 2 & 2 & 2 & 2 & 2 \\
    & {tps} & 32 & 32 & 32 & 32 & 32 & 32 & 32 & 32 & 32 \\
    & {tpf} & 16 & 16 & 16 & 16 & 16 & 16 & 16 & 16 & 16 \\
    \midrule 
    \multirow{3}{*}{\rotatebox[origin=c]{90}{\footnotesize \textit{Data}}}
    & \textbf{Dataset} & A & A & A & VpT& VpT   & VpT  & SFT & SFT & SFT \\
    & \hspace{3mm}\#Samples & 198K & 198K & 198K & 396K & 396K & 396K & 3.2M & 3.2M & 3.2M \\
    & \hspace{3mm}Type & I+V & I+V  & I+V  & V & V & V & T+I+MI+V & T+I+MI+V & T+I+MI+V \\

    \midrule
    \multirow{4}{*}{\rotatebox[origin=c]{90}{\footnotesize \textit{Model}}}
    & \textbf{Trainable} &  38.4M & 63.6M  & 177M &  1.4B  & 1.5B   & 1.6B   &   1.6B & 3.2B   & 7.8B   \\
    & \hspace{3mm}{$\psi_{\text{vision}}$} & -- &  --   & -- &  1.4B  & 1.4B   & 1.4B   &  -- & --   & --   \\
    & \hspace{3mm}{$\theta_{\text{connector}}$} & 38.4M & 63.6M  & 177M & 38.4M & 63.6M  & 177M  & 38.4M & 63.6M  & 177M \\
    & \hspace{3mm}{$\phi_{\text{LLM}}$} & -- & --   & --  & -- & --   & -- & 1.54B & 3.09B  & 7.62B \\
    \midrule 
    \multirow{5}{*}{\rotatebox[origin=c]{90}{\footnotesize \textit{Training}}}
    & \textbf{Batch Size} & 256 & 256 & 256 & 256 & 256 & 256 & 256 & 256 & 256 \\
    & \textbf{LR: $\psi_{\text{vision}}$} & 0 & 0 & 0 & 5 $\times 10^{-6}$ & 5 $\times 10^{-6}$ & 5 $\times 10^{-6}$ & 0 & 0 & 0 \\    
    & \textbf{LR: $\theta_{\text{connector}}$} & 1$\times 10^{-4}$ & 1$\times 10^{-4}$ & 1$\times 10^{-4}$ & 1 $\times 10^{-4}$ & 1 $\times 10^{-4}$ & 1 $\times 10^{-4}$ & 1 $\times 10^{-4}$ & 1 $\times 10^{-4}$ & 1 $\times 10^{-4}$ \\
    & \textbf{LR: $\phi_{\text{LLM}}$} &  0 & 0 & 0 &  0 & 0 & 0 & 5 $\times 10^{-5}$ & 2 $\times 10^{-5}$ & 1 $\times 10^{-5}$ \\
    & \textbf{Epoch} & 1 & 1 & 1 & 1 & 1 & 1 & 1 & 1 & 1 \\
    \bottomrule
    \end{NiceTabular}
    }\vspace{-0.05in}
    \caption{\textbf{Detailed configuration for each training stage of \method.} The table summarizes the maximum clips per video, frames per second (fps), dataset information, trainable parameters, and training hyperparameters across different stages of training (\textbf{Alignment}, \textbf{Vision pretraining}, \textbf{SFT}) for \method models of varying sizes (1.5B, 3B, and 7.6B).}
    \label{tab:hyperparameters}
\end{table*}

\section{\method implementation details}\label{app:sec:implementation}
In this section, we provide detailed descriptions of all the design decisions in \method, including implementation specifics, hyperparameters, and other relevant details.

\subsection{Architecture}

\method encodes clips consisting of $N$ frames, where $N$ depends on the video encoder used ($=4$ for InternVideo2+SigLIP-SO400M). 
We opted for a fully shared pipeline for both images and videos, so when encoding images, we replicate the image $N$ times to match the clip length.
Video frames are then encoded independently with the InternVideo2 and SigLIP-SO400M encoders.
The output features are interpolated and concatenated along the channel dimension before being fed to a connector module. The connector projects the features to match the hidden dimension of the Large Language Model, and the resampler resamples them into a predetermined number of $T$ tokens per clip using the Perciver Resampler. An overview of \method is shown in Fig.~\ref{fig:apollo_arch}. For vision-text token integration, we utilize the \texttt{clip from \{MM:SS\}-\{MM:SS\}: <vid\_token>}.

\method effectively samples videos as a series of independent clips. By keeping the clip sampling frames per second (fps) constant, the model learns to reason about fine-grained temporal aspects, such as the speed of objects. Many previous methods employ uniform frame sampling, especially when handling long videos, effectively changing the ``playback speed'' between iterations. In contrast, we sample clips uniformly spaced throughout the video, and if the video is too long, we distribute the individual clips uniformly rather than adjusting the frame sampling rate. We, therefore, sample clips concurrently until reaching the maximum number of clips (see Tab.~\ref{tab:hyperparameters}), at which point we start uniformly distancing the clips.

\subsection{Unified vs. Split Architectures}
\label{sec:training:unified_split}

While previous sections focused on different aspects of design and training protocols, we also investigated the impact of using a unified versus a split architecture for integrating image and video modalities.
A unified architecture processes both image and video inputs through the same set of encoders and token resamplers, ensuring a single consistent visual representation path. In contrast, a split architecture separates the processing streams for images and videos, potentially offering more specialized representations at the cost of increased complexity.
Previously,~\cite{video_llava} advocated for sharing the mlp connector between images and videos, claiming that this leads to better transfer.~\cite{chat_univi} performed the same token merging and utilized the same connector for both images and videos. Recent works encode video frames entirely independently, completely removing the need for separate architectures for image and video inputs.

As shown in Tab.~\ref{table:split_vs_unified}, our experiments revealed that the unified architecture performs slightly better or on par with the split architecture across key benchmarks. The unified approach strikes an appealing balance between performance and simplicity, offering a more elegant and parameter-efficient solution. Given these findings, we adopt the unified architecture as our default setting for \method.
\begin{table}[ht]
    \centering
    \setlength\tabcolsep{1.9mm}
    \adjustbox{width=0.65\textwidth}{
    \begin{NiceTabular}{lcccccc}
    \CodeBefore
    \rectanglecolor{metabg}{2-7}{4-7}
    \Body
    \toprule
    Archi- & \multicolumn{6}{c}{\benchmark} \\
     \cmidrule(lr){2-7}
    tecture& OCR & Spatial & Egocentric & Perception & Reasoning & Overall \\
    \midrule
    Split  & 46.2 & 55.7 & 62.3 & 59.0 & 58.1 & 56.2 \\
    Unified  & 50.0 & 54.0 & 61.7 & 60.8 & 57.9 &\textbf{56.8} \\
    \bottomrule
    \end{NiceTabular}
    }\caption{
    \textbf{Split vs Unified Architectures on \benchmark.} 
    A comparison of the performance across different tasks, including OCR, Spatial, Egocentric, Perception, and Reasoning, as well as the overall score.
    \label{table:split_vs_unified}
    }
\end{table}

\subsection{Data}\label{sup:method:data}
We utilized a diverse mix of publicly available and licensed datasets across text, image-text, multi-image, and video modalities. Due to licensing restrictions, we excluded non-permissive datasets—such as those leveraging ChatGPT—which limited our inclusion of some commonly used datasets. We generated multi-turn conversations to enrich our training data by leveraging Large Multimodal Models (LMMs), such as Qwen2VL-7B, for captioning. Then, we used LLaMA 3.1 70B~\citep{touvron2023llama} to convert these captions into conversations.
Detailed data statistics are presented in Fig.~\ref{fig:data_mixture}. It is possible that performance could be further improved without such restrictions and by training on larger datasets like those introduced in LLaVA-OneVision~\cite{llava} and Cambrian1~\cite{cambrian1}.

Our training process comprised three distinct stages:
\begin{enumerate} 
\item \textbf{Alignment}: In this phase, we trained on a 198K mixture of 50/50 image and video captions.
\item \textbf{Vision Pretraining}: We tuned the encoders using a video-only caption dataset of 396K samples. 
\item \textbf{Supervised Fine-tuning (SFT)}: We trained on a mixture of text, image, multi-image, and video data, with a total of 3.2 million samples. 
\end{enumerate}

\subsection{Training}
We trained our models using 128 NVIDIA A100 GPUs. Due to the large-scale nature of this study, we automated model training to be spawned from csv files, which would automatically update with the final evaluations. Most experiments were done with ZeRO2 optimization, as full model sharding was unnecessary for our models, but ZeRO3 is supported for future researchers interested in training larger models. 
We utilized the AdamW optimizer for all training stages with a gradient clipping threshold of 1. We applied a warm-up ratio of 0.03 and a cosine learning rate schedule. The training objective was the cross-entropy loss for autoregressive text generation only.
We adjusted the learning rates of the Large Language Model (LLM) components proportionally to the square root of their relative model sizes. We found that employing a higher learning rate for the connector module yielded the best performance.

\section{\designscale: efficient model design with smaller models}
\label{app:sec:scaling}

Developing Large Multi-modal Models (LMMs) with billions of parameters is computationally intensive. A key question is whether smaller models can reliably inform design decisions for larger ones. We introduce \designscale, a phenomenon where design choices evaluated on moderately sized models (approximately 2–4 billion parameters) correlate highly with those on larger models, enabling efficient model development.

To investigate \designscale, we conducted extensive experiments varying key aspects of LMM design, such as architecture, video sampling, training strategies, and data mixtures. We selected $21$ distinct model variations encompassing these design dimensions. Each variation was trained using four different Large Language Models: Qwen2-0.5B, Qwen2-1.5B, Qwen1.5-4B, and Qwen2-7B, resulting in a total of 84 models.

Unlike traditional scaling laws—which typically require training multiple models from within the same model family to understand how performance scales with size—\designscale allows us to transfer design insights without such extensive efforts. In scaling laws, researchers train around 3–5 models of different sizes to establish scaling relationships, and only then can they determine which design decisions are beneficial at larger scales. 
In contrast, \designscale shows that design decisions on moderately sized models transfer well to larger ones, even across different model families. Our primary goal is to show that design decisions transfer reliably, reducing computational burden and accelerating research.

\begin{figure}[!ht]
\centering
\includegraphics[width=0.9\linewidth]{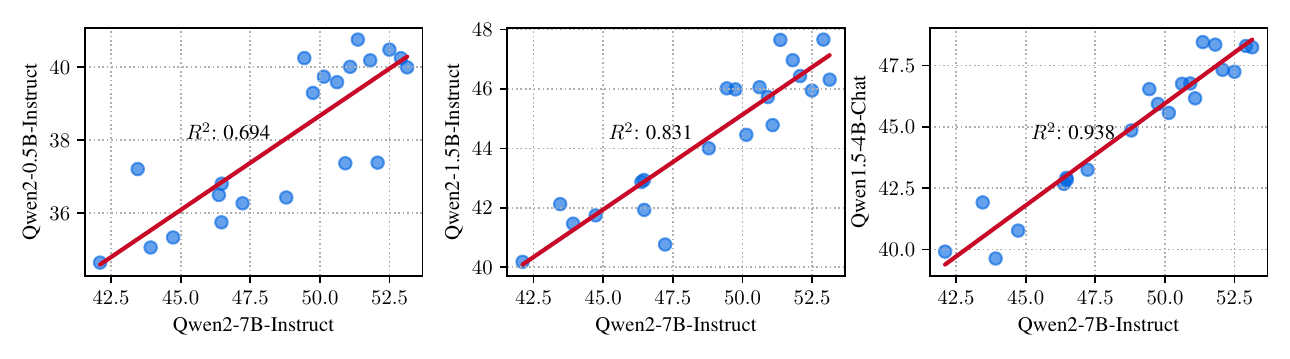}\vspace{-0.1in}
\includegraphics[width=0.9\linewidth]{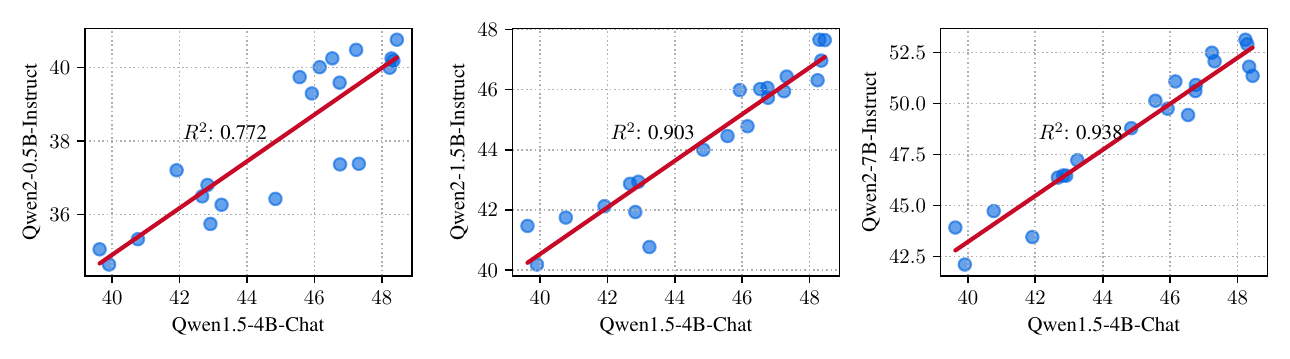}\vspace{-0.1in}
\includegraphics[width=0.9\linewidth]{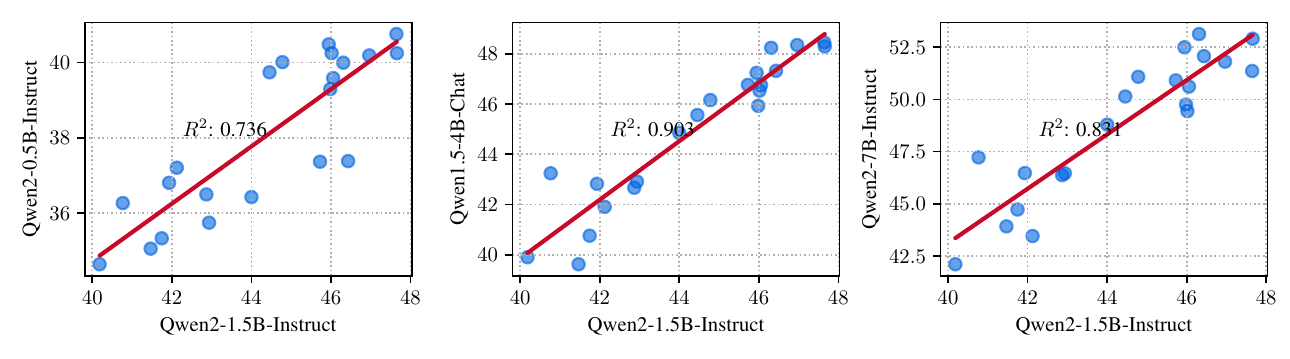}\vspace{-0.1in}
\includegraphics[width=0.9\linewidth]{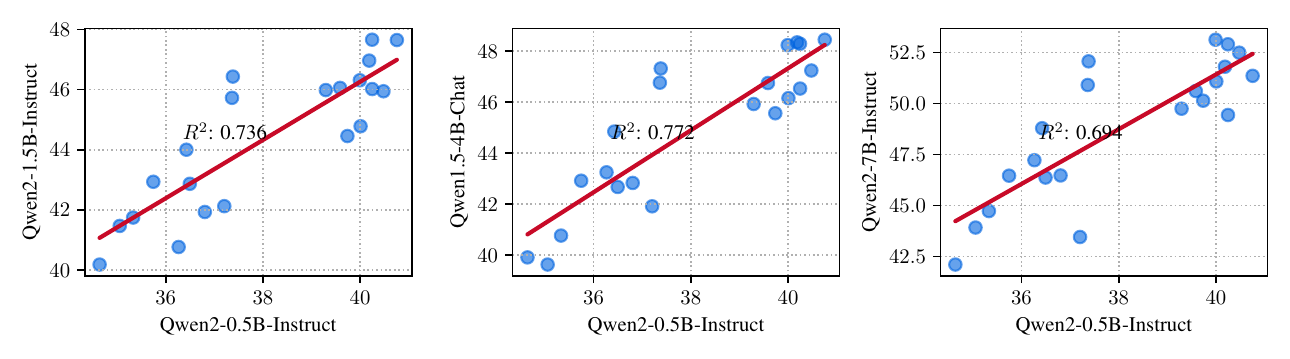}\vspace{-0.1in}
\caption{\label{sup:fig:scaling_consistency} \textbf{\designscale.} Average accuracy for each one for each design variation, 
we can tell model's correlation gets progressively better. When comparing two small models (1.5B and 0.5B), we do not see a good correlation, confirming that the \designscale is not due to the models being of similar size but larger than a certain size. }
\end{figure}

In Fig.~\ref{sup:fig:scaling_consistency}, we present all the correlation plots from our study. When comparing the 7B model to smaller ones (first row), we observe that the $R^2$ progressively increases with model size. A similar pattern is seen when comparing the 4B model to smaller models. For the 1.5B model, however, the $R^2$ decreases when compared to larger models, and with the 0.5B model, the $R^2$ is essentially random.
We find that the $R^2$ behaves log-linearly with model size. This suggests that at around 3 billion parameters, we can expect an $R^2$ greater than 0.9 when compared with the 7B model. Since the behavior is log-linear, models above the 3–4 billion parameter range can be expected to have high correlation even with much larger models, such as 32B ($>R^2\simeq0.86$) or 72B parameters ($>R^2\simeq0.84$).

\section{Effect of video sampling on the different dimensions of video perception}
\label{app:sec:sampling}

Fig.~\ref{sup:fig:full_sampling} presents a detailed analysis of how varying frames per second (fps) and tokens per second (tps) impact our model's performance across different video perception tasks: Optical Character Recognition (OCR), Spatial Understanding, Egocentric Understanding, Perception, and Reasoning. Our findings indicate that OCR and Spatial Understanding tasks exhibit a uniform and steep decline in performance when tps is reduced, particularly noticeable at lower values of 2--4 tps, regardless of fps settings. This suggests that these tasks are highly sensitive to the amount of visual information encoded per frame, significantly affecting performance by the number of tokens per frame.

In contrast, Egocentric Understanding and Reasoning tasks show a less severe performance drop when tps is reduced, especially at lower fps values. This implies that these tasks are less sensitive to the number of tokens per frame and are more influenced by the temporal resolution provided by fps, with the ability to capture temporal dynamics being more critical than the density of visual information per frame. The Perception metric behaves as an outlier; apart from an anomalous data point at 1~fps, perception performance tends to favor lower fps values and is less affected by variations in tps. This indicates that for certain perceptual tasks, higher temporal sampling does not necessarily provide additional benefits, and effective performance can be achieved with fewer frames and tokens.

Overall, these results highlight the importance of tailoring video sampling strategies to the specific requirements of different video perception tasks to optimize model performance.

\section{Raw results}
\label{app:sec:raw}
We provide the raw evaluations of all the models utilized in our study. Many investigations required multiple experiments to test whether design decisions hold under multiple hyperparameters. 
We provide all the raw data used in our study for further analysis. 
    For Sec.~\ref{section:scaling-laws}: Tab.~\ref{tab:results:raw_scaling1} \&~\ref{tab:results:raw_scaling2},
    Sec.~\ref{sec:arch:sampling}: Tab.~\ref{tab:fps_sampling_raw} \&~\ref{tab:uniform_sampling_raw},
    Sec.~\ref{sec:arch:encoders}: Tab~\ref{tab:vision_encoder_raw},
    Sec.~\ref{sec:arch:integration}: Tab.~\ref{table:temp_embed}, 
    Sec.~\ref{sec:training:stages}: Tab.~\ref{table:training_stages_raw}, 
    Sec.~\ref{sec:training:data}: Tab.~\ref{tab:data_mix_raw}.

\FloatBarrier

\begin{table*}
\centering
    \resizebox{\textwidth}{!}{%
\begin{NiceTabular}{l l l c c c c c c}
    \CodeBefore
    \rectanglecolor{metabg}{2-9}{-9}
    \Body
\toprule
& \multicolumn{2}{c}{Hyperparameters} & \multicolumn{6}{c}{\benchmark} \\
\cmidrule(lr){2-3} \cmidrule(lr){4-9}
 &                 LLM &                Vision Encoders &  OCR & Spatial & Egocentric & Perception & Reasoning & Overall \\
\midrule
1 & Qwen2.5-3B-Instruct &                         DINOv2 & 36.6 &    40.5 &       55.9 &       48.0 &      46.3 &    45.5 \\
2 & Qwen2.5-3B-Instruct &                 LanguageBind-Image & 41.2 &    46.2 &       49.5 &       51.0 &      51.7 &    47.9 \\
3 & Qwen2.5-3B-Instruct &                  SigLIP SO400M & 41.9 &    52.2 &       57.4 &       52.0 &      60.0 &    52.7 \\
4 & Qwen2.5-3B-Instruct &                       VideoMAE & 35.6 &    35.5 &       47.9 &       47.0 &      40.0 &    41.2 \\
5 & Qwen2.5-3B-Instruct &                         V-JEPA & 39.4 &    35.2 &       44.1 &       52.0 &      44.6 &    43.1 \\
6 & Qwen2.5-3B-Instruct &                 LanguageBind-Video & 41.2 &    47.8 &       54.8 &       53.2 &      46.3 &    48.7 \\
7 & Qwen2.5-3B-Instruct &                   InternVideo2 & 43.7 &    46.5 &       56.4 &       55.2 &      58.1 &    52.0 \\
8 & Qwen2.5-3B-Instruct &               VideoMAE + DINOv2 & 40.1 &    43.2 &       57.4 &       59.5 &      47.5 &    49.6 \\
9 & Qwen2.5-3B-Instruct &       VideoMAE + LanguageBind-Image & 39.8 &    49.8 &       55.9 &       57.5 &      49.8 &    50.5 \\
10 & Qwen2.5-3B-Instruct &        VideoMAE + SigLIP SO400M & 45.8 &    54.8 &       55.9 &       63.0 &      55.6 &    55.0 \\
11 & Qwen2.5-3B-Instruct &                 V-JEPA + DINOv2 & 41.5 &    43.2 &       56.4 &       55.2 &      48.5 &    49.0 \\
12 & Qwen2.5-3B-Instruct &         V-JEPA + LanguageBind-Image & 43.3 &    49.2 &       50.5 &       59.2 &      52.9 &    51.1 \\
13 & Qwen2.5-3B-Instruct &          V-JEPA + SigLIP SO400M & 48.6 &    53.2 &       59.0 &       57.8 &      58.1 &    55.3 \\
14 & Qwen2.5-3B-Instruct & LanguageBind-Video + DINOv2 & 41.5 & 44.9 & 54.6 & 57.6 & 51.0 & 50.0\\
15 & Qwen2.5-3B-Instruct & LanguageBind-Video + LanguageBind-Image & 41.2 &    48.5 &       53.2 &       62.7 &      54.7 &    52.1 \\
16 & Qwen2.5-3B-Instruct &  LanguageBind-Video + SigLIP SO400M & 45.4 &    50.5 &       59.6 &       56.8 &      54.9 &    53.4 \\
17 & Qwen2.5-3B-Instruct &           InternVideo2 + DINOv2 & 43.0 &    48.2 &       50.0 &       58.0 &      57.1 &    51.3 \\
18 & Qwen2.5-3B-Instruct &   InternVideo2 + LanguageBind-Image & 45.8 &    48.0 &       51.6 &       62.3 &      56.9 &    52.9 \\
19 & Qwen2.5-3B-Instruct &    InternVideo2 + SigLIP SO400M & 48.8 &    56.4 &       59.9 &       64.1 &      64.5 &    57.9 \\

\bottomrule
\end{NiceTabular}
}
\caption{\textbf{Raw results for vision encoders experiment.} The table presents performance scores on \benchmark at a tokens-per-second (TPS) rate of 32. Metrics include OCR, spatial understanding, egocentric reasoning, perception, reasoning, and overall performance. The encoders are grouped and ordered as follows: single image encoders, single video encoders, and dual encoder configurations.}
\label{tab:vision_encoder_raw}
\end{table*}

\begin{table*}
\centering
    \resizebox{\textwidth}{!}{%
\begin{NiceTabular}{l l l c c c c c c c c c}
    \CodeBefore
    \rectanglecolor{metabg}{2-12}{30-12}
    \Body
\toprule
    & \multicolumn{5}{c}{Hyperparameters} & \multicolumn{6}{c}{\benchmark} \\
\cmidrule(lr){2-6} \cmidrule(lr){7-12}
 &                 LLM &                          Vision Encoders &   tps & fps &   tpf &  OCR & Spatial & Egocentric & Perception & Reasoning & Overall \\
\midrule
  1 & Qwen2.5-3B-Instruct & InternVideo2-1B + SigLIP SO400M & 512.0 & 4.0 & 128.0 & 46.0 &    51.0 &       52.1 &       59.0 &      54.0 &    52.4 \\
  2 & Qwen2.5-3B-Instruct & InternVideo2-1B + SigLIP SO400M & 256.0 & 2.0 & 128.0 & 45.5 &    53.5 &       51.5 &       59.0 &      49.0 &    51.7 \\
  3 & Qwen2.5-3B-Instruct & InternVideo2-1B + SigLIP SO400M & 128.0 & 1.0 & 128.0 & 51.0 &    55.0 &       55.3 &       51.0 &      62.5 &    54.9 \\
  4 & Qwen2.5-3B-Instruct & InternVideo2-1B + SigLIP SO400M &  64.0 & 0.5 & 128.0 & 48.0 &    52.0 &       54.2 &       63.0 &      56.0 &    54.6 \\
  5 & Qwen2.5-3B-Instruct & InternVideo2-1B + SigLIP SO400M & 256.0 & 4.0 &  64.0 & 43.5 &    50.0 &       55.3 &       62.0 &      56.0 &    53.3 \\
  6 & Qwen2.5-3B-Instruct & InternVideo2-1B + SigLIP SO400M & 128.0 & 2.0 &  64.0 & 51.0 &    52.0 &       61.6 &       58.0 &      59.5 &    56.4 \\
  7 & Qwen2.5-3B-Instruct & InternVideo2-1B + SigLIP SO400M &  64.0 & 1.0 &  64.0 & 52.5 &    55.0 &       60.6 &       58.5 &      57.0 &    56.7 \\
  8 & Qwen2.5-3B-Instruct & InternVideo2-1B + SigLIP SO400M &  32.0 & 0.5 &  64.0 & 47.5 &    56.5 &       60.0 &       58.0 &      60.0 &    56.4 \\
  9 & Qwen2.5-3B-Instruct & InternVideo2-1B + SigLIP SO400M & 128.0 & 4.0 &  32.0 & 52.0 &    57.5 &       60.6 &       61.0 &      57.5 &    57.7 \\
 10 & Qwen2.5-3B-Instruct & InternVideo2-1B + SigLIP SO400M &  64.0 & 2.0 &  32.0 & 55.0 &    58.0 &       60.6 &       55.5 &      62.5 &    58.3 \\
 11 & Qwen2.5-3B-Instruct & InternVideo2-1B + SigLIP SO400M &  32.0 & 1.0 &  32.0 & 52.5 &    54.5 &       62.7 &       51.0 &      63.0 &    56.7 \\
 12 & Qwen2.5-3B-Instruct & InternVideo2-1B + SigLIP SO400M &  16.0 & 0.5 &  32.0 & 50.0 &    56.0 &       58.4 &       63.0 &      58.0 &    57.1 \\
 13 & Qwen2.5-3B-Instruct & InternVideo2-1B + SigLIP SO400M &  64.0 & 4.0 &  16.0 & 49.5 &    60.5 &       58.4 &       60.0 &      62.5 &    58.2 \\
 14 & Qwen2.5-3B-Instruct & InternVideo2-1B + SigLIP SO400M &  32.0 & 2.0 &  16.0 & 53.0 &    56.0 &       53.1 &       56.0 &      59.5 &    55.6 \\
 15 & Qwen2.5-3B-Instruct & InternVideo2-1B + SigLIP SO400M &  16.0 & 1.0 &  16.0 & 54.5 &    58.5 &       55.3 &       61.0 &      61.0 &    58.1 \\
 16 & Qwen2.5-3B-Instruct & InternVideo2-1B + SigLIP SO400M &   8.0 & 0.5 &  16.0 & 50.0 &    50.5 &       61.1 &       59.5 &      55.5 &    55.3 \\
 17 & Qwen2.5-3B-Instruct & InternVideo2-1B + SigLIP SO400M &  32.0 & 4.0 &   8.0 & 55.5 &    59.5 &       59.0 &       57.5 &      61.5 &    58.6 \\
 18 & Qwen2.5-3B-Instruct & InternVideo2-1B + SigLIP SO400M &  16.0 & 2.0 &   8.0 & 45.5 &    55.5 &       60.0 &       66.0 &      62.5 &    57.9 \\
 19 & Qwen2.5-3B-Instruct & InternVideo2-1B + SigLIP SO400M &   8.0 & 1.0 &   8.0 & 54.5 &    55.0 &       62.7 &       59.0 &      58.0 &    57.8 \\
 20 & Qwen2.5-3B-Instruct & InternVideo2-1B + SigLIP SO400M &   4.0 & 0.5 &   8.0 & 50.5 &    56.0 &       57.4 &       61.5 &      60.0 &    57.1 \\  
 21 & Qwen2.5-3B-Instruct & InternVideo2-1B + SigLIP SO400M &  16.0 & 4.0 &   4.0 & 29.5 &    25.0 &        1.0 &       38.5 &      12.5 &    21.5 \\
 22 & Qwen2.5-3B-Instruct & InternVideo2-1B + SigLIP SO400M &   8.0 & 2.0 &   4.0 & 35.0 &    40.5 &       48.9 &       52.0 &      40.0 &    43.2 \\
 23 & Qwen2.5-3B-Instruct & InternVideo2-1B + SigLIP SO400M &   4.0 & 1.0 &   4.0 & 41.5 &    43.5 &       52.1 &       63.0 &      51.5 &    50.3 \\
 24 & Qwen2.5-3B-Instruct & InternVideo2-1B + SigLIP SO400M &   2.0 & 0.5 &   4.0 & 39.5 &    47.0 &       61.6 &       55.0 &      50.0 &    50.5 \\
 25 & Qwen2.5-3B-Instruct & InternVideo2-1B + SigLIP SO400M &   8.0 & 4.0 &   2.0 & 38.5 &    36.5 &       54.2 &       47.5 &      44.5 &    44.1 \\
 26 & Qwen2.5-3B-Instruct & InternVideo2-1B + SigLIP SO400M &   4.0 & 2.0 &   2.0 & 26.5 &    23.5 &       30.8 &       44.9 &      27.5 &    32.4 \\
 27 & Qwen2.5-3B-Instruct & InternVideo2-1B + SigLIP SO400M &   2.0 & 1.0 &   2.0 & 37.3 &    41.8 &       53.1 &       50.3 &      45.9 &    44.8 \\
 28 & Qwen2.5-3B-Instruct & InternVideo2-1B + SigLIP SO400M &   1.0 & 0.5 &   2.0 & 41.0 &    42.0 &       54.2 &       45.0 &      48.2 &    47.3 \\
 \bottomrule
\end{NiceTabular}
}
\caption{
\textbf{Raw results of video sampling experiment.} \benchmark breaks down metrics to OCR, spatial understanding, egocentric reasoning, perception, reasoning, and overall performance. The table highlights the impact of frames per second (fps), tokens per second (tps), and tokens per frame (tpf).}
\label{tab:fps_sampling_raw}
\end{table*}

\begin{table*}
\centering
    \resizebox{\textwidth}{!}{%
\begin{NiceTabular}{lllcccccccc}
    \CodeBefore
    \rectanglecolor{metabg}{2-11}{11-11}
    \Body
\toprule
    & \multicolumn{4}{c}{Hyperparameters} & \multicolumn{6}{c}{\benchmark} \\
\cmidrule(lr){2-5} \cmidrule(lr){6-11}
& \multirow{2}{*}{LLM} & \multirow{2}{*}{Vision Encoders} & \multicolumn{2}{c}{Uniform Frames} & \multirow{2}{*}{OCR} & \multirow{2}{*}{Spatial} & \multirow{2}{*}{Egocentric} & \multirow{2}{*}{Perception} & \multirow{2}{*}{Reasoning} & \multirow{2}{*}{Overall} \\

&                      &                                  & \makecell{(Train)} & \makecell{(Test)} &                         &                         &                             &                             &                             &                         \\

\midrule
  1 & Qwen2.5-3B-Instruct & InternVideo2-1B + SigLIP SO400M &                      8 &                     8 & 38.0 &    41.0 &       43.1 &       50.3 &      44.0 &    44.2 \\
  2 & Qwen2.5-3B-Instruct & InternVideo2-1B + SigLIP SO400M &                     16 &                    16 & 40.5 &    46.7 &       55.9 &       55.3 &      46.1 &    48.1 \\
  3 & Qwen2.5-3B-Instruct & InternVideo2-1B + SigLIP SO400M &                     32 &                    32 & 49.5 &    52.0 &       51.1 &       58.5 &      48.5 &    51.9 \\
  4 & Qwen2.5-3B-Instruct & InternVideo2-1B + SigLIP SO400M &                     64 &                    64 & 46.5 &    52.0 &       61.2 &       56.5 &      59.5 &    55.1 \\
\midrule
  5 & Qwen2.5-3B-Instruct & InternVideo2-1B + SigLIP SO400M &                      8 &                    No & 42.5 &    44.5 &       54.8 &       52.0 &      51.5 &    49.0 \\
  6 & Qwen2.5-3B-Instruct & InternVideo2-1B + SigLIP SO400M &                     16 &                    No & 48.0 &    43.5 &       58.5 &       60.5 &      53.0 &    52.6 \\
  7 & Qwen2.5-3B-Instruct & InternVideo2-1B + SigLIP SO400M &                     32 &                    No & 46.0 &    50.0 &       52.1 &       57.5 &      57.5 &    52.6 \\
  8 & Qwen2.5-3B-Instruct & InternVideo2-1B + SigLIP SO400M &                     64 &                    No & 48.5 &    53.5 &       59.0 &       54.5 &      54.0 &    53.8 \\
\bottomrule
\end{NiceTabular}
}
\caption{
\textbf{Raw results of uniform sampling experiment.} \benchmark evaluates metrics including OCR, spatial understanding, egocentric reasoning, perception, reasoning, and overall performance. Top half are results when models are both trained and tested with uniform frame sampling. The bottom half is when the models are trained with uniform frame sampling but tested at an fps of 2.}
\label{tab:uniform_sampling_raw}
\end{table*}

\begin{table*}[t]
    \centering
    \adjustbox{width=\textwidth}{
    \begin{NiceTabular}{lccccccccc}
    \CodeBefore
    \rectanglecolor{metabg}{2-10}{19-10}
    \Body
    \toprule
    & \multicolumn{3}{c}{Training Stages} & \multicolumn{6}{c}{\benchmark} \\
    \cmidrule(lr){2-4} \cmidrule(lr){5-10}
    & \lmmicon &  \lmmicon &  \lmmicon  & OCR & Spatial& Egocentric & Perception & Reasoning & Overall \\
    \midrule
         1 & $0,1e^{-4},3e^{-5}$ & - & -       &  42.0 & 46.5& 54.9 & 50.0 & 49.5 & 48.7  \\ %
         2& $1e^{-6},1e^{-4},3e^{-5}$  & - & - & 28.8 & 29.2 & 18.8 & 35.5 & 22.6 & 30.8 \\ %
         3& $5e^{-6},1e^{-4},3e^{-5}$  & - & - & 26.8 & 23.2 & 12.1 & 21.4 & 24.5 & 22.2 \\
         4& $1e^{-5},1e^{-4},3e^{-5}$  & - & - & 24.9 & 16.1 & 26.4 & 39.9 & 18.4 & 25.0  \\
        \midrule 
         5& $0,1e^{-4},0,$ & $0,1e^{-4},3e^{-5}$ & -              & 52.2 & 54.5& 55.9 & 60.3 & 58.4 & 56.3 \\ %
         6& $1e^{-6},1e^{-4},0$  & $0,1e^{-4},3e^{-5}$ & -        & 49.6 & 54.2 & 61.4 & 63.3 & 59.5 & 57.6 \\ %
         7& $5e^{-6},1e^{-4},0$  & $0,1e^{-4},3e^{-5}$ & -        &51.6 & 54.5 & 58.0& 62.1 & 60.2 & 57.8  \\ %
         8& $1e^{-5},1e^{-4},0$  & $0,1e^{-4},3e^{-5}$ & -        & 50.0 & 50.0 & 44.5 & 55.3 & 47.6 & 49.7 \\ %
         9& $1e^{-6},1e^{-4},0$  & $1e^{-6},1e^{-4},3e^{-5}$ & -  & 42.2 & 48.9 & 61.7 & 43.7 & 52.2 & 48.1 \\ %
         10& $5e^{-6},1e^{-4},0$  & $5e^{-6},1e^{-4},3e^{-5}$ & - & 32.2 & 37.5 & 50.0 & 40.4 & 44.2 & 40.3  \\ %
         11& $1e^{-5},1e^{-4},0$  & $1e^{-5},1e^{-4},3e^{-5}$ & - & 30.3 & 23.8 & 49.5 & 41.9 & 30.3 & 33.7 \\ %
    \midrule 
         12& $0,1e^{-4},0,$  & $1e^{-6},1e^{-4},0$ & $0,1e^{-4},3e^{-5}$   & 46.7 & 50.7 & 60.6 & 57.6 & 61.8 & 55.4 \\ %
         13& $0,1e^{-4},0,$  & $5e^{-6},1e^{-4},0$ & $0,1e^{-4},3e^{-5}$   & 52.4 & 55.4 & 62.8 & 63.5 & 61.4 & 59.2  \\%
         14& $0,1e^{-4},0,$  & $1e^{-5},1e^{-4},0$ & $0,1e^{-4},3e^{-5}$   & 53.7 & 54.0 & 47.3 & 56.2 & 52.9 & 53.2 \\ %

         15& $0,1e^{-4},0,$  & $1e^{-6},1e^{-4},0$ & $1e^{-6},1e^{-4},3e^{-5}$   & 44.2 & 37.5 & 43.9 & 56.6 & 38.5 & 44.2  \\ %
         16& $0,1e^{-4},0,$  & $5e^{-6},1e^{-4},0$ & $5e-5,1e^{-4},3e^{-5}$      & 32.7 & 36.9 & 49.8 & 40.1 & 44.5 & 39.8  \\ %
         17& $0,1e^{-4},0,$  & $1e^{-5},1e^{-4},0$ & $1e^{-5},1e^{-4},3e^{-5}$   & 32.4 & 36.6 & 30.1 & 42.3 & 33.5 & 35.4 \\ %
    \bottomrule
    \end{NiceTabular}
    }\vspace{-0.1in}
    \caption{
    \textbf{Raw results of training schedules experiments.} Results of training models across 1, 2, and 3 stages with varying learning rates (LR) and data mixtures. The table highlights OCR, spatial understanding, egocentric reasoning, perception, reasoning, and overall performance metrics. Each stage utilizes different LR configurations and data distributions, showing the benefits of multi-stage training for optimizing performance across all metrics (see Tab.~\ref{tab:hyperparameters} and Sec.~\ref{sup:method:data} for details).
    }
    \label{table:training_stages_raw}
\end{table*}

\begin{table*}
\setlength\tabcolsep{3.5mm}
\centering
    \resizebox{\textwidth}{!}{%
\begin{NiceTabular}{lcccccccccc}
    \CodeBefore
    \rectanglecolor{metabg}{2-11}{15-11}
    \Body
\toprule
    & \multicolumn{4}{c}{Data Composition} & \multicolumn{6}{c}{\benchmark} \\
\cmidrule(lr){2-5} \cmidrule(lr){6-11}
& {Text} & {Image} & {Multi-Image} & {Video} & {OCR} & {Spatial} & {Egocentric} & {Perception} & {Reasoning} & {Overall} \\
\midrule
   1 &  25.0 &   25.0 &         25.0 &   25.0 & 41.0 &     49.5 &        59.0 &        57.0 &       59.5 &     54.1 \\
   2 &  15.0 &   25.0 &         20.0 &   40.0 & 47.5 &     59.0 &        60.6 &        66.0 &       62.0 &     59.0 \\
   3 &  15.0 &   32.5 &         20.0 &   32.5 & 46.5 &     52.0 &        58.0 &        65.5 &       63.5 &     57.1 \\
   4 &  15.0 &   40.0 &         20.0 &   25.0 & 45.0 &     57.3 &        52.1 &        60.2 &       61.1 &     56.2 \\
   5 &   7.0 &   38.7 &         20.0 &   34.3 & 44.5 &     53.0 &        54.3 &        58.0 &       55.5 &     53.0 \\
   6 &   7.0 &   55.0 &         20.0 &   18.0 & 39.5 &     45.0 &        46.7 &        56.0 &       54.0 &     48.3 \\
   7 &   7.0 &  18.0 &         48.0 &   27.0 &  40.2 &     48.0 &        53.2 &        57.2 &       53.5 &     50.9 \\
   8 &   7.0 &    0.0 &         0.0 &   93.0 & 37.5 &     33.5 &        52.7 &        40.5 &       45.5 &     41.8 \\
   9 &   7.0 &    0.0 &         20.0 &   73.0 & 37.0 &     44.0 &        51.1 &        45.0 &       49.0 &     45.1 \\
   10 &   7.0 &   14.0 &         18.0 &   61.0 & 41.0 &     48.5 &        54.2 &        56.8 &       54.5 &     51.2  \\
   11 &  5.0 &   10.0 &         40.0 &   45.0 & 40.0 &     47.5 &        53.2 &        57.3 &       51.7 &     50.4 \\
   12 &  2.0 &   30.0 &         30.0 &   38.0 & 35.5 &     47.0 &        55.0 &        56.0 &       49.5 &     48.7 \\
   13 &  0.0 &   38.7 &         20.0 &   41.3 & 35.4 &     44.1 &        54.1 &        54.2 &       49.0 &     47.5 \\
\bottomrule
\end{NiceTabular}
}
\caption{\label{tab:data_mix_raw}
    \textbf{Raw results of data composition experiments.} Performance outcomes of video-based Large Multi-modal Models (LMMs) trained with varying proportions of Text, Image, Multi-Image, and Video data mixtures. The table presents benchmark scores across OCR, Spatial, Egocentric, Perception, Reasoning, and Overall performance metrics for each distinct data composition. These results emphasize the critical role of balanced data mixtures in optimizing model performance (see Sec.~\ref{sec:training:data} for details).
}
\end{table*}

\begin{table*}
\centering
    \resizebox{\textwidth}{!}{%
\begin{tabular}{l l l c c c c c c c c}
        \toprule
         & 
        \textbf{LLM} & 
        \textbf{Vision Towers} & 
        \makecell{\textbf{Vision} \\ \textbf{Freeze}} & 
        \makecell{\textbf{Clip} \\ \textbf{Duration}} & 
        \makecell{\textbf{Tokens} \\ \textbf{/Clip}} & 
        \textbf{fps} & 
        \textbf{tps} & 
        \makecell{\textbf{Tokens} \\ \textbf{/Frame}} & 
        \makecell{\textbf{Data} \\ \textbf{Mixture}} & 
        \textbf{Average} \\
        \midrule
1 & Qwen2-7B-Instruct & LanguageBind-Video-v1.5 + SigLIP SO400M & \freeze & 5 & 32 & 1.6 & 6.4 & 4 & A & 46.37 \\
2 & Qwen2-7B-Instruct & LanguageBind-Video-v1.5 + SigLIP SO400M & \freeze & 5 & 64 & 1.6 & 12.8 & 8 & A & 46.46\\
3 & Qwen2-7B-Instruct & LanguageBind-Video-v1.5 + SigLIP SO400M & \train & 5 & 64 & 1.6 & 12.8 & 8 & A & 48.79 \\
4 & Qwen2-7B-Instruct & VJEPA-H@384 + SigLIP SO400M & \freeze & 5 & 32 & 3.2 & 6.4 & 2 & A & 47.22 \\
5 & Qwen2-7B-Instruct & VJEPA-H@384 + SigLIP SO400M & \freeze & 10 & 64 & 1.6 & 6.4 & 4 & A & 43.46 \\
6 & Qwen2-7B-Instruct & VJEPA-H@384 + SigLIP SO400M & \freeze & 5 & 64 & 3.2 & 12.8 & 2 & A & 46.47 \\
7 & Qwen2-7B-Instruct & VJEPA-H@384 + SigLIP SO400M & \train & 5 & 64 & 3.2 & 12.8 & 2 & A & 42.11 \\
8 & Qwen2-7B-Instruct & LanguageBind-Video-v1.5 + SigLIP SO400M & \freeze & 5 & 32 & 1.6 & 6.4 & 4 & B & 49.75 \\
9 & Qwen2-7B-Instruct & LanguageBind-Video-v1.5 + SigLIP SO400M & \freeze & 5 & 64 & 1.6 & 12.8 & 8 & B & 50.61 \\
10 & Qwen2-7B-Instruct & LanguageBind-Video-v1.5 + SigLIP SO400M & \train & 5 & 64 & 1.6 & 12.8 & 8 & B & 50.91 \\
11 & Qwen2-7B-Instruct & VJEPA-H@384 + SigLIP SO400M & \freeze & 5 & 32 & 3.2 & 6.4 & 2 & B & 49.44 \\
12 & Qwen2-7B-Instruct & VJEPA-H@384 + SigLIP SO400M & \freeze & 10 & 64 & 1.6 & 6.4 & 4 & B & 50.14 \\
13 & Qwen2-7B-Instruct & VJEPA-H@384 + SigLIP SO400M & \freeze & 5 & 64 & 3.2 & 12.8 & 2 & B & 51.08 \\
14 & Qwen2-7B-Instruct & VJEPA-H@384 + SigLIP SO400M & \train & 5 & 64 & 3.2 & 12.8 & 2 & B & 43.92 \\
15 & Qwen2-7B-Instruct & LanguageBind-Video-v1.5 + SigLIP SO400M & \freeze & 5 & 32 & 1.6 & 6.4 & 4 & C & 51.80 \\
16 & Qwen2-7B-Instruct & LanguageBind-Video-v1.5 + SigLIP SO400M & \freeze & 5 & 64 & 1.6 & 12.8 & 8 & C & 52.91 \\
17 & Qwen2-7B-Instruct & LanguageBind-Video-v1.5 + SigLIP SO400M & \train & 5 & 64 & 1.6 & 12.8 & 8 & C & 52.07 \\
18 & Qwen2-7B-Instruct & VJEPA-H@384 + SigLIP SO400M & \freeze & 5 & 32 & 3.2 & 6.4 & 2 & C & 51.36 \\
19 & Qwen2-7B-Instruct & VJEPA-H@384 + SigLIP SO400M & \freeze & 10 & 64 & 1.6 & 6.4 & 4 & C & 52.49 \\
20 & Qwen2-7B-Instruct & VJEPA-H@384 + SigLIP SO400M & \freeze & 5 & 64 & 3.2 & 12.8 & 2 & C & 53.13 \\
21 & Qwen2-7B-Instruct & VJEPA-H@384 + SigLIP SO400M & \train & 5 & 64 & 3.2 & 12.8 & 2 & C & 44.73 \\
22 & Qwen1.5-4B-Chat & LanguageBind-Video-v1.5 + SigLIP SO400M & \freeze & 5 & 32 & 1.6 & 6.4 & 4 & A & 42.67 \\
23 & Qwen1.5-4B-Chat & LanguageBind-Video-v1.5 + SigLIP SO400M & \freeze & 5 & 64 & 1.6 & 12.8 & 8 & A & 42.92 \\
24 & Qwen1.5-4B-Chat & LanguageBind-Video-v1.5 + SigLIP SO400M & \train & 5 & 64 & 1.6 & 12.8 & 8 & A & 44.85 \\
25 & Qwen1.5-4B-Chat & VJEPA-H@384 + SigLIP SO400M & \freeze & 5 & 32 & 3.2 & 6.4 & 2 & A & 43.25 \\
26 & Qwen1.5-4B-Chat & VJEPA-H@384 + SigLIP SO400M & \freeze & 10 & 64 & 1.6 & 6.4 & 4 & A & 41.91 \\
27 & Qwen1.5-4B-Chat & VJEPA-H@384 + SigLIP SO400M & \freeze & 5 & 64 & 3.2 & 12.8 & 2 & A & 42.83 \\
28 & Qwen1.5-4B-Chat & VJEPA-H@384 + SigLIP SO400M & \train & 5 & 64 & 3.2 & 12.8 & 2 & A & 39.91 \\
29 & Qwen1.5-4B-Chat & LanguageBind-Video-v1.5 + SigLIP SO400M & \freeze & 5 & 32 & 1.6 & 6.4 & 4 & B & 45.90 \\
30 & Qwen1.5-4B-Chat & LanguageBind-Video-v1.5 + SigLIP SO400M & \freeze & 5 & 64 & 1.6 & 12.8 & 8 & B & 46.75 \\
31 & Qwen1.5-4B-Chat & LanguageBind-Video-v1.5 + SigLIP SO400M & \train & 5 & 64 & 1.6 & 12.8 & 8 & B & 46.76 \\
32 & Qwen1.5-4B-Chat & VJEPA-H@384 + SigLIP SO400M & \freeze & 5 & 32 & 3.2 & 6.4 & 2 & B & 46.53 \\
33 & Qwen1.5-4B-Chat & VJEPA-H@384 + SigLIP SO400M & \freeze & 10 & 64 & 1.6 & 6.4 & 4 & B & 45.56 \\
34 & Qwen1.5-4B-Chat & VJEPA-H@384 + SigLIP SO400M & \freeze & 5 & 64 & 3.2 & 12.8 & 2 & B & 46.16 \\
35 & Qwen1.5-4B-Chat & VJEPA-H@384 + SigLIP SO400M & \train & 5 & 64 & 3.2 & 12.8 & 2 & B & 39.63 \\
36 & Qwen1.5-4B-Chat & LanguageBind-Video-v1.5 + SigLIP SO400M & \freeze & 5 & 32 & 1.6 & 6.4 & 4 & C & 48.34 \\
37 & Qwen1.5-4B-Chat & LanguageBind-Video-v1.5 + SigLIP SO400M & \freeze & 5 & 64 & 1.6 & 12.8 & 8 & C & 48.29 \\
38 & Qwen1.5-4B-Chat & LanguageBind-Video-v1.5 + SigLIP SO400M & \train & 5 & 64 & 1.6 & 12.8 & 8 & C & 47.32 \\
39 & Qwen1.5-4B-Chat & VJEPA-H@384 + SigLIP SO400M & \freeze & 5 & 32 & 3.2 & 6.4 & 2 & C & 48.45 \\
40 & Qwen1.5-4B-Chat & VJEPA-H@384 + SigLIP SO400M & \freeze & 10 & 64 & 1.6 & 6.4 & 4 & C & 47.24 \\
41 & Qwen1.5-4B-Chat & VJEPA-H@384 + SigLIP SO400M & \freeze & 5 & 64 & 3.2 & 12.8 & 2 & C & 48.24 \\
42 & Qwen1.5-4B-Chat & VJEPA-H@384 + SigLIP SO400M & \train & 5 & 64 & 3.2 & 12.8 & 2 & C & 40.76 \\

\bottomrule
\end{tabular}}
\caption{\textbf{Raw results of \designscale experiments (1/2).} This table presents the raw performance data of 42 model configurations used in the \designscale experiments. Each configuration explores the effect of various parameters, including the LLM size (Qwen variants), vision tower configurations, freezing or training vision encoders, clip duration, tokens per clip, frames per second (fps), tokens per second (tps), tokens per frame, and data mixture. The ``Average'' column reports the overall performance score. These results support the investigation into how smaller models can serve as proxies for larger models in determining effective design decisions.\label{tab:results:raw_scaling1}}

\end{table*}

\begin{table*}
\centering
    \resizebox{\textwidth}{!}{%
\begin{tabular}{l l l c c c c c c c c}
        \toprule
         & 
        \textbf{LLM} & 
        \textbf{Vision Towers} & 
        \makecell{\textbf{Vision} \\ \textbf{Freeze}} & 
        \makecell{\textbf{Clip} \\ \textbf{Duration}} & 
        \makecell{\textbf{Tokens} \\ \textbf{/Clip}} & 
        \textbf{fps} & 
        \textbf{tps} & 
        \makecell{\textbf{Tokens} \\ \textbf{/Frame}} & 
        \makecell{\textbf{Data} \\ \textbf{Mixture}} & 
        \textbf{Average} \\
        \midrule
             
43 & Qwen2-1.5B-Instruct & LanguageBind-Video-v1.5 + SigLIP SO400M & \freeze & 5 & 32 & 1.6 & 6.4 & 4 & A & 42.87 \\
44 & Qwen2-1.5B-Instruct & LanguageBind-Video-v1.5 + SigLIP SO400M & \freeze & 5 & 64 & 1.6 & 12.8 & 8 & A & 42.94 \\
45 & Qwen2-1.5B-Instruct & LanguageBind-Video-v1.5 + SigLIP SO400M & \train & 5 & 64 & 1.6 & 12.8 & 8 & A & 44.00 \\
46 & Qwen2-1.5B-Instruct & VJEPA-H@384 + SigLIP SO400M & \freeze & 5 & 32 & 3.2 & 6.4 & 2 & A & 40.77 \\
47 & Qwen2-1.5B-Instruct & VJEPA-H@384 + SigLIP SO400M & \freeze & 10 & 64 & 1.6 & 6.4 & 4 & A & 42.13 \\
48 & Qwen2-1.5B-Instruct & VJEPA-H@384 + SigLIP SO400M & \freeze & 5 & 64 & 3.2 & 12.8 & 2 & A & 41.93 \\
49 & Qwen2-1.5B-Instruct & VJEPA-H@384 + SigLIP SO400M & \train & 5 & 64 & 3.2 & 12.8 & 2 & A & 40.18 \\
50 & Qwen2-1.5B-Instruct & LanguageBind-Video-v1.5 + SigLIP SO400M & \freeze & 5 & 32 & 1.6 & 6.4 & 4 & B & 45.98 \\
51 & Qwen2-1.5B-Instruct & LanguageBind-Video-v1.5 + SigLIP SO400M & \freeze & 5 & 64 & 1.6 & 12.8 & 8 & B & 46.06 \\
52 & Qwen2-1.5B-Instruct & LanguageBind-Video-v1.5 + SigLIP SO400M & \train & 5 & 64 & 1.6 & 12.8 & 8 & B & 45.73 \\
53 & Qwen2-1.5B-Instruct & VJEPA-H@384 + SigLIP SO400M & \freeze & 5 & 32 & 3.2 & 6.4 & 2 & B & 46.02 \\
54 & Qwen2-1.5B-Instruct & VJEPA-H@384 + SigLIP SO400M & \freeze & 10 & 64 & 1.6 & 6.4 & 4 & B & 44.46 \\
55 & Qwen2-1.5B-Instruct & VJEPA-H@384 + SigLIP SO400M & \freeze & 5 & 64 & 3.2 & 12.8 & 2 & B & 44.78 \\
56 & Qwen2-1.5B-Instruct & VJEPA-H@384 + SigLIP SO400M & \train & 5 & 64 & 3.2 & 12.8 & 2 & B & 41.47 \\
57 & Qwen2-1.5B-Instruct & LanguageBind-Video-v1.5 + SigLIP SO400M & \freeze & 5 & 32 & 1.6 & 6.4 & 4 & C & 46.96 \\
58 & Qwen2-1.5B-Instruct & LanguageBind-Video-v1.5 + SigLIP SO400M & \freeze & 5 & 64 & 1.6 & 12.8 & 8 & C & 47.66 \\
59 & Qwen2-1.5B-Instruct & LanguageBind-Video-v1.5 + SigLIP SO400M & \train & 5 & 64 & 1.6 & 12.8 & 8 & C & 46.43 \\
60 & Qwen2-1.5B-Instruct & VJEPA-H@384 + SigLIP SO400M & \freeze & 5 & 32 & 3.2 & 6.4 & 2 & C & 47.65 \\
61 & Qwen2-1.5B-Instruct & VJEPA-H@384 + SigLIP SO400M & \freeze & 10 & 64 & 1.6 & 6.4 & 4 & C & 45.94 \\
62 & Qwen2-1.5B-Instruct & VJEPA-H@384 + SigLIP SO400M & \freeze & 5 & 64 & 3.2 & 12.8 & 2 & C & 46.31 \\
63 & Qwen2-1.5B-Instruct & VJEPA-H@384 + SigLIP SO400M & \train & 5 & 64 & 3.2 & 12.8 & 2 & C & 41.76 \\
64 & Qwen2-0.5B-Instruct & LanguageBind-Video-v1.5 + SigLIP SO400M & \freeze & 5 & 32 & 1.6 & 6.4 & 4 & A & 36.50 \\
65 & Qwen2-0.5B-Instruct & LanguageBind-Video-v1.5 + SigLIP SO400M & \freeze & 5 & 64 & 1.6 & 12.8 & 8 & A & 35.75 \\
66 & Qwen2-0.5B-Instruct & LanguageBind-Video-v1.5 + SigLIP SO400M & \train & 5 & 64 & 1.6 & 12.8 & 8 & A & 36.43 \\
67 & Qwen2-0.5B-Instruct & VJEPA-H@384 + SigLIP SO400M & \freeze & 5 & 32 & 3.2 & 6.4 & 2 & A & 36.27 \\
68 & Qwen2-0.5B-Instruct & VJEPA-H@384 + SigLIP SO400M & \freeze & 10 & 64 & 1.6 & 6.4 & 4 & A & 37.21 \\
69 & Qwen2-0.5B-Instruct & VJEPA-H@384 + SigLIP SO400M & \freeze & 5 & 64 & 3.2 & 12.8 & 2 & A & 36.80 \\
70 & Qwen2-0.5B-Instruct & VJEPA-H@384 + SigLIP SO400M & \train & 5 & 64 & 3.2 & 12.8 & 2 & A & 34.64 \\
71 & Qwen2-0.5B-Instruct & LanguageBind-Video-v1.5 + SigLIP SO400M & \freeze & 5 & 32 & 1.6 & 6.4 & 4 & B & 39.29 \\
72 & Qwen2-0.5B-Instruct & LanguageBind-Video-v1.5 + SigLIP SO400M & \freeze & 5 & 64 & 1.6 & 12.8 & 8 & B & 39.59 \\
73 & Qwen2-0.5B-Instruct & LanguageBind-Video-v1.5 + SigLIP SO400M & \train & 5 & 64 & 1.6 & 12.8 & 8 & B & 37.36 \\
74 & Qwen2-0.5B-Instruct & VJEPA-H@384 + SigLIP SO400M & \freeze & 5 & 32 & 3.2 & 6.4 & 2 & B & 40.25 \\
75 & Qwen2-0.5B-Instruct & VJEPA-H@384 + SigLIP SO400M & \freeze & 10 & 64 & 1.6 & 6.4 & 4 & B & 39.74 \\
76 & Qwen2-0.5B-Instruct & VJEPA-H@384 + SigLIP SO400M & \freeze & 5 & 64 & 3.2 & 12.8 & 2 & B & 40.01 \\
77 & Qwen2-0.5B-Instruct & VJEPA-H@384 + SigLIP SO400M & \train & 5 & 64 & 3.2 & 12.8 & 2 & B & 35.05 \\
78 & Qwen2-0.5B-Instruct & LanguageBind-Video-v1.5 + SigLIP SO400M & \freeze & 5 & 32 & 1.6 & 6.4 & 4 & C & 40.19 \\
79 & Qwen2-0.5B-Instruct & LanguageBind-Video-v1.5 + SigLIP SO400M & \freeze & 5 & 64 & 1.6 & 12.8 & 8 & C & 40.25 \\
80 & Qwen2-0.5B-Instruct & LanguageBind-Video-v1.5 + SigLIP SO400M & \train & 5 & 64 & 1.6 & 12.8 & 8 & C & 37.38 \\
81 & Qwen2-0.5B-Instruct & VJEPA-H@384 + SigLIP SO400M & \freeze & 5 & 32 & 3.2 & 6.4 & 2 & C & 40.76 \\
82 & Qwen2-0.5B-Instruct & VJEPA-H@384 + SigLIP SO400M & \freeze & 10 & 64 & 1.6 & 6.4 & 4 & C & 40.48 \\
83 & Qwen2-0.5B-Instruct & VJEPA-H@384 + SigLIP SO400M & \freeze & 5 & 64 & 3.2 & 12.8 & 2 & C & 39.99 \\
84 & Qwen2-0.5B-Instruct & VJEPA-H@384 + SigLIP SO400M & \train & 5 & 64 & 3.2 & 12.8 & 2 & C & 35.33 \\ 
\bottomrule
\end{tabular}
}
\caption{\textbf{Raw results of \designscale experiments (2/2).} This table presents the raw performance data of 42 model configurations used in the \designscale experiments. Each configuration explores the effect of various parameters, including the LLM size (Qwen variants), vision tower configurations, freezing or training vision encoders, clip duration, tokens per clip, frames per second (fps), tokens per second (tps), tokens per frame, and data mixture. The ``Average'' column reports the overall performance score. These results support the investigation into how smaller models can serve as proxies for larger models in determining effective design decisions.\label{tab:results:raw_scaling2}}
\end{table*}

\end{document}